\documentclass[10pt,twocolumn,letterpaper]{article}

\usepackage[pagenumbers]{cvpr} % To force page numbers, e.g. for an arXiv version
\usepackage{algorithm}
\usepackage{makecell}  
\usepackage{amsmath}
\usepackage{gensymb}
\usepackage{multirow}
\definecolor{cvprblue}{rgb}{0.21,0.49,0.74}
\usepackage[pagebackref,breaklinks,colorlinks,allcolors=cvprblue]{hyperref}
\usepackage[accsupp]{axessibility} 

\def\paperID{7666} % *** Enter the Paper ID here
\def\confName{CVPR}
\def\confYear{2026}

\title{Generalizable and Relightable Gaussian Splatting for \\Human Novel View Synthesis}

\author{Yipengjing Sun$^1$, Shengping Zhang$^{*, 1}$, Chenyang Wang$^1$, Shunyuan Zheng$^1$, Zonglin Li$^1$, Xiangyang Ji$^2$\\
$^1$Harbin Institute of Technology
$^2$Tsinghua University\\
{\tt\small \{yipengjing.sun, c.wang\}@stu.hit.edu.cn, \{sawyer0503, zonglin.li, s.zhang\}@hit.edu.cn}\\
{\tt\small xyji@tsinghua.edu.cn
}
}

\begin{document}

\twocolumn[{
\maketitle
\begin{figure}[H]
\hsize=\textwidth %
\centering
\vspace{-14mm}
\includegraphics[width=0.96\textwidth]{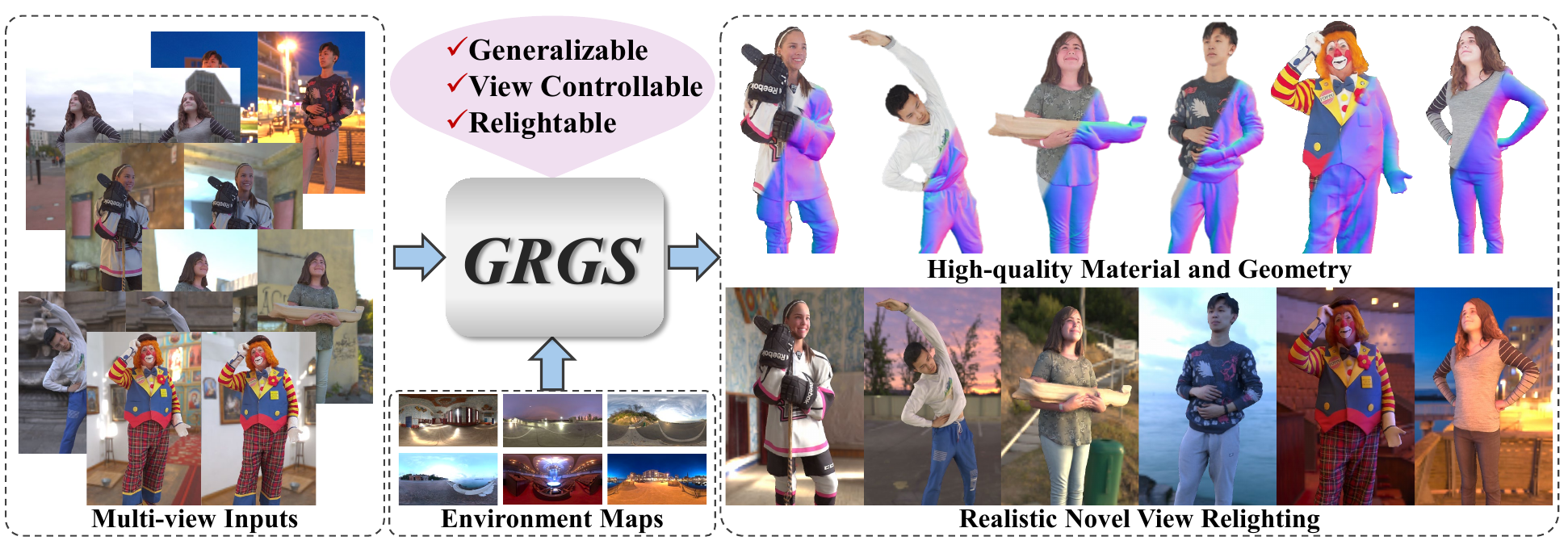}
\vspace{-3mm}
\caption{Given multi-view inputs and environment maps, GRGS reconstructs generalizable 3D representations with high-quality geometry and material while supporting realistic relighting rendering from arbitrary viewpoints.}
\label{fig:teaser}
% \vspace{-3mm}
\end{figure}
}]

% \maketitle

% \begin{figure*}[h]
% % \vspace{-8mm}
% \hsize=\textwidth %
% \centering

% \includegraphics[width=\textwidth]{figs/teaser_new_v3.pdf}
% % \vspace{-6mm}
% \caption{Given multi-view inputs and environment maps, GRGS reconstructs generalizable 3D representations with high-quality geometry and material while supporting realistic relighting rendering from arbitrary viewpoints. } 
% \vspace{-2mm}
% \label{fig:teaser}
% \end{figure*}
\let\thefootnote\relax\footnotetext{$^*$ Corresponding author.}
\begin{abstract}
We propose GRGS, a generalizable and relightable 3D Gaussian framework for high-fidelity human novel view synthesis under diverse lighting conditions.
Unlike existing methods that rely on per-character optimization or ignore physical constraints, GRGS adopts a feed-forward, fully supervised strategy projecting geometry, material, and illumination cues from multi-view 2D observations into 3D Gaussian representations.
To recover accurate geometry under diverse lighting conditions, we introduce a Lighting-robust Geometry Refinement (LGR) module trained on synthetically relit data to predict precise depth and surface normals.
Based on the high-quality geometry, a Physically Grounded Neural Rendering (PGNR) module is further proposed to integrate neural prediction with physics-based shading, supporting editable relighting with shadows and indirect illumination.
Moreover, we design a 2D-to-3D projection training scheme leveraging differentiable supervision from ambient occlusion, direct, and indirect lighting maps, alleviating the computational cost of ray tracing.
Extensive experiments demonstrate that GRGS achieves superior visual quality, geometric consistency, and generalization across characters and lighting conditions.
The project webpage is available at {\small\url{https://sypj-98.github.io/grgs/}}.
% We propose GRGS, a generalizable and relightable 3D Gaussian framework for high-fidelity human novel view synthesis under diverse lighting conditions.
% Unlike existing methods that rely on per-character optimization or ignore physical constraints, GRGS adopts a feed-forward, fully supervised strategy that projects geometry, material, and illumination cues from multi-view 2D observations into 3D Gaussian representations.
% % Specifically, to reconstruct lighting-invariant geometry,
% Specifically, to recover accurate geometry under diverse lighting conditions, we introduce a Lighting-robust Geometry Refinement (LGR) module trained on synthetically relit data to predict accurate depth and surface normals.
% Based on the high-quality geometry, a Physically Grounded Neural Rendering (PGNR) module is further proposed to integrate neural prediction with physics-based shading, supporting editable relighting with shadows and indirect illumination.
% Besides, we design a 2D-to-3D projection training scheme that leverages differentiable supervision from ambient occlusion, direct, and indirect lighting maps, which alleviates the computational cost of explicit ray tracing.
% Extensive experiments demonstrate that GRGS achieves superior visual quality, geometric consistency, and generalization across characters and lighting conditions. 
\end{abstract}    
\section{Introduction}
\label{sec:intro}

Human novel view synthesis (NVS) aims to produce photorealistic images of human performers under a specific targeting novel viewpoint, which has a wide range of applications such as immersive telepresence, cinematic production, and AR/VR. 
To further improve realism and fidelity, relightable rendering, which enables editing lighting during synthesizing novel views, has become an important improvement over traditional NVS, as lighting significantly affects surface appearance and immersive shading.
Although significant efforts have been devoted to developing effective relightable rendering, it is still a challenging task due to complex light transport in 3D space.

Prior methods~\cite{debevec2012light,li2013capturing,imber2014intrinsic, guo2019relightables} typically rely on mesh-based geometry and texture reconstruction to support plausible viewpoint transitions, but they often involve complex reconstruction pipelines and struggle with generating accurate and smooth mesh surfaces, limiting their effectiveness for realistic relighting without extensive post-processing.
Recent advances~\cite{Lombardi:2019, mildenhall2021nerf, kerbl20233d} in neural representations have substantially propelled the field of NVS. 
Specifically, NeRF-based approaches~\cite{srinivasan2021nerv, zhang2021nerfactor, jin2023tensoir, yao2022neilf, li2024tensosdf, chen2022relighting4d, xu2024relightable} incorporate physically-based rendering (PBR) principles into volumetric radiance fields or signed distance fields (SDFs) for implicit geometry and material encoding, enabling the synthesis of photorealistic images under novel viewpoints and lighting conditions.
However, these methods typically require time-consuming per-character optimization and incur high computational costs, resulting in limited rendering speeds that hinder their practicality in real-world applications.
Recently, 3D Gaussian Splatting (3DGS)~\cite{kerbl20233d} has emerged as an efficient and explicit neural representation, offering a fast training and inference process while maintaining high visual fidelity. 
To adapt 3DGS for relighting tasks, researchers~\cite{gao2024relightable, jiang2024gaussianshader, liang2024gs, wu2025deferredgs, li2024animatable} augment each 3D Gaussian point with intrinsic properties (geometry and appearance) and employ inverse rendering techniques to estimate light transport. 
%
% Although this strategy has achieved promising near real-time relighting performance, it remains reliant on iterative inverse rendering pipelines that are inherently ill-posed. 
% %
% Consequently, the estimated geometry, materials, and lighting often lack physical accuracy, degrading rendering quality.
Unfortunately, it remains reliant on iterative inverse rendering pipelines that are inherently ill-posed, resulting in inaccurate estimation of geometry and materials, which degrades the overall rendering quality.
Moreover, per-character optimization still prevents them from being applied to relighting applications that require generalization across different characters.
% the absence of a feed-forward inference mechanism continues to hinder their practical deployment.

In contrast to 3D methods, 2D image-based relighting approaches employ encoder-decoder architectures (\textit{e.g.}, U-Net)~\cite{pandey2021total,  ji2022geometry, mei2023lightpainter, kim2024switchlight} or diffusion models~\cite{ponglertnapakorn2023difareli, ren2024relightful, zhang2025scaling} to learn lighting priors from large-scale relighting datasets, producing high-quality relit images under novel illumination in a generalized manner. 
Despite their efficiency and strong generalization capabilities, the lack of 3D consistent constraints makes 2D image-based methods suffer from flickering artifacts when rendering from precisely user-controlled viewpoints.
In addition, although these methods often produce visually impressive relighting results, they often overlook physical interpretability as they entirely rely on neural networks to implicitly capture physical constraints.
% they often ignore physical interpretability due to their reliance on neural networks to implicitly learn physical constraints.

To address these challenges, we propose GRGS, as shown in Fig.~\ref{fig:teaser}, a generalizable and relightable 3D Gaussian framework for high-fidelity human novel view synthesis under various lighting conditions by integrating multiple intrinsic attribute priors into 3D Gaussian representations.
Unlike existing 3DGS-based methods that apply person-specific optimization, the core idea of GRGS is to adopt a supervised, data-driven strategy that learns to project geometry, material, and illumination cues from multi-view 2D observations onto 3D Gaussian attributes in a feed-forward manner for robust generalization and realistic relighting.
Our framework starts by presenting a Lighting-robust Geometry Refinement (LGR) module, which first estimates coarse depth using a stereo-based method and then refines per-Gaussian surface normals to capture smooth and fine-grained details.
To obtain reliable geometry under challenging illumination, we synthesize large-scale relit multi-view data, enabling LGR to produce geometry consistent across diverse lighting conditions, thereby mitigating feature mismatches and improving geometric accuracy.
% enabling LGR to produce lighting-invariant geometry that mitigates feature mismatches and improves geometric accuracy.
The refined geometric priors not only support realistic relighting but also serve as a critical bridge between the 2D image space and the 3D Gaussian domain, facilitating accurate positioning of Gaussians.
With the geometry established, we design a Physically Grounded Neural Rendering (PGNR) module that integrates physics-based rendering with neural rendering, which synthesizes realistic shading phenomena, including shadows and indirect lighting, while enforcing physically grounded lighting consistency.
PGNR employs geometry-aware decoders to infer Gaussian parameters and intrinsic attributes from the refined geometry in a feed-forward manner.
In parallel, a lightweight encoder-decoder processes the high-resolution environment map to estimate direct illumination scaling factors and spherical harmonic coefficients for modeling indirect lighting.
By incorporating a physically-based rendering, our framework ensures physically consistent light transport, achieving high-quality and photorealistic relighting results.
Besides, the 2D-to-3D project strategy alleviates the computational cost of explicit ray tracing for visibility and global illumination, ensuring high efficiency during inference.
In summary, our method makes the following key contributions:
% \begin{figure}[t]
% % \vspace{-7pt}
% 	\centering
% 	\includegraphics[width=\textwidth]{figs/teaser_new.pdf}
% 	% \vspace{-7pt}
% 	\caption{}
% 	\label{fig:overveiw}
%  % \vspace{-2pt}
% \end{figure}
\begin{itemize}
    \item We propose GRGS, a generalizable and relightable 3D Gaussian framework that projects geometry, material, and illumination cues from multi-view 2D observations onto 3D Gaussian attributes in a feed-forward manner, enabling realistic and robust novel view synthesis of unseen data under novel lighting conditions.
    \item We present a Lighting-roubst Geometry Refinement module trained on synthetically relit multi-view data to estimate accurate depth and surface normals, effectively mitigating geometry errors caused by uneven illumination.
    \item We design a Physically Grounded Neural Rendering module that combines physics-based rendering with neural rendering, achieving high-quality shading effects with improved computational efficiency.
    % We design a Physically Grounded Neural Rendering module that combines physics-based rendering with neural rendering, supporting high-quality shading phenomena synthesis while avoiding the computational overhead of explicit ray tracing.
    % \item We introduce a depth-guided normal refinement module that addresses surface detail loss and mitigates grid-like artifacts commonly observed when computing normals from binocular depth estimates, thereby significantly enhancing geometric fidelity for relighting.
\end{itemize}
\section{Related Work}
\label{sec:formatting}

\paragraph{Person-specific human relighting.}
Traditional methods~\cite{debevec2000acquiring, hawkins2001photometric, debevec2002lighting, wenger2005performance, debevec2012light, weyrich2006analysis, chabert2006relighting, guo2019relightables} propose to sample the reflectance field of a human performer through a LightStage setup comprising controlled lighting systems and dense camera arrays, which can generate photorealistic renderings under novel lighting environments. 
However, the costly setup and complex person-specific relighting process constrain their widespread applications. 
With the rapid progress in neural implicit representations~\cite{mildenhall2021nerf, aliev2020neural, bi2020deep, lombardi2019neural, thies2019deferred, xu2019deep, xu2018deep}, neural inverse rendering~\cite{boss2021nerd, srinivasan2021nerv, zhang2021nerfactor, zhang2021physg, boss2021neural, jin2023tensoir, yao2022neilf, zhang2023neilf++} has emerged as a promising approach for person-specific human relighting, enabling the joint recovery of geometry, material, and illumination from multi-view images captured under arbitrary lighting conditions.
Recent variants~\cite{chen2022relighting4d, sun2023neural, xu2023relightable, lin2024relightable, li2024animatable, chen2024meshavatar, wang2024intrinsicavatar} extend inverse rendering to dynamic performers by integrating a parametric human body template into the learning of implicit fields, enabling temporally coherent reconstruction under varying poses and motions.
However, these methods typically require time-consuming person-specific optimization and incur high computational costs, resulting in limited training and rendering speeds.
With the advent of explicit point-based 3DGS~\cite{kerbl20233d} representation, recent 3DGS-based inverse rendering methods~\cite{gao2024relightable, jiang2024gaussianshader, liang2024gs, wu2025deferredgs, hong2025beam} explore the disentanglement of material, geometry, and lighting through the use of Gaussian points, enabling efficient training and fast inference.
ARGS~\cite{li2024animatable} pioneers the integration of animatable 3D Gaussians and inverse rendering, delivering visually compelling and relightable full-body human avatars.
RGCA~\cite{saito2024relightable} leverages dynamic relighting data captured in a LightStage and an explicit mesh template to achieve high-quality relightable head avatars. However, such methods typically rely on time-consuming person-specific optimization. 
Moreover, performing inverse rendering from multi-view images under arbitrary illumination is inherently ill-posed, making it particularly difficult to accurately disentangle geometry, material, and lighting, ultimately limiting the fidelity of the final relighting results.

\paragraph{Generalizable human relighting.}
With the advancement of deep learning, data-driven methods~\cite{pandey2021total, sun2019single, zhou2019deep, wang2020single, mei2023lightpainter} have achieved impressive results in portrait relighting from a single image, primarily by leveraging convolutional neural networks trained on synthetic data derived from OLAT datasets. However, these methods typically overlook underlying human geometry, leading to physically implausible light–shadow interactions. Recent efforts~\cite{meka2020deep, ji2022geometry} attempt to address this limitation by estimating human geometry, but monocular predictions remain insufficiently accurate for reliable relighting. Other approaches~\cite{kanamori2019relighting, lagunas2021single, tajima2021relighting} attempt to learn light transport directly from 2D shading images. However, modeling complex 3D light transport in 2D space remains fundamentally challenging. Such methods are generally restricted to approximating low-frequency lighting effects and fail to generalize to unseen illumination. SwitchLight~\cite{kim2024switchlight} circumvents the need for explicit 3D geometry by introducing physics-based rendering in the image domain, achieving high-quality relighting results. Nevertheless, the absence of an explicit 3D representation limits its ability to handle occlusions and maintain spatial consistency. More recently, diffusion-based generative models~\cite{ponglertnapakorn2023difareli, ren2024relightful, zhang2025scaling} demonstrate impressive relighting performance by leveraging strong pre-trained priors and high-quality lighting datasets, producing visually striking results with high generalization ability. However, priors learned from pre-training and Light Stage supervision lack physics-based constraints during training, which in turn reduces the physical interpretability and realism of the outputs.
% However, the learned priors from pre-training and the lighting supervision often lack physically grounded constraints, limiting the physical interpretability and realism of the generated outputs.
\section{Method}
\begin{figure*}[t]
\vspace{-15pt}
	\centering
	\includegraphics[width=\textwidth]{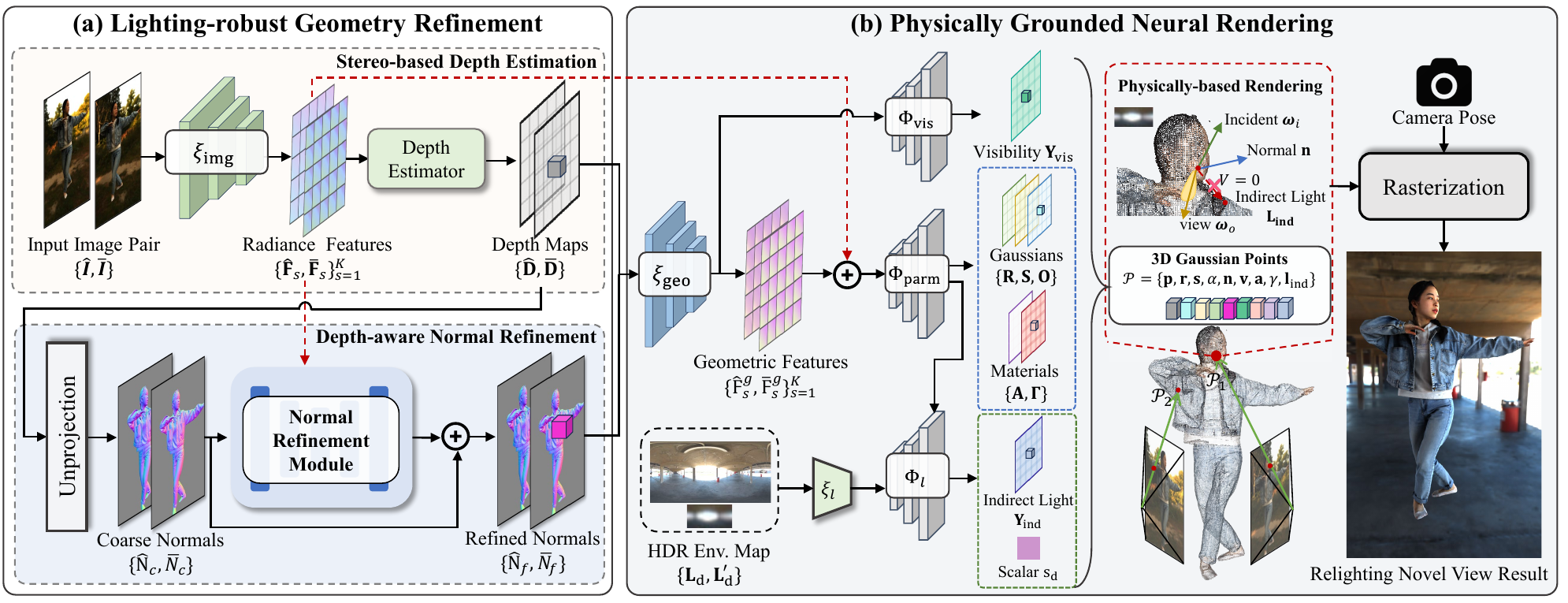}
	\vspace{-15pt}
	\caption{\textbf{Overview of GRGS.} Given sparse-view images of a performer under arbitrary illumination, GRGS first leverages the LGR module to reconstruct accurate depth and surface normals, and then employs the PGNR module for material decomposition and physically plausible realistic relighting from novel viewpoints. }
	\label{fig:pipeline}
 \vspace{-15pt}
\end{figure*}

As illustrated in Fig.~\ref{fig:pipeline}, our proposed GRGS achieves generalizable and relightable human novel view synthesis via two core modules: (1) Lighting-robust Geometry Refinement (LGR) module and (2) Physically Grounded Neural Rendering (PGNR) module.
The core idea is to adopt a supervised, data-driven strategy that projects intrinsic attributes from multi-view 2D observations onto 3D Gaussian representations in a feed-forward manner.
Specifically, given a set of sparse-view human images, GRGS first constructs the LGR module to estimate depth using a stereo-based method and then refines per-Gaussian surface normals via a depth-aware refinement, which allows GRGS to mitigate geometry errors caused by uneven illumination (Sec.~\ref{sec: depth estimation}).
Next, building on this accurate geometry, the PGNR module fuses physics-based rendering with neural rendering to synthesize realistic relighting phenomena.
It leverages geometry-aware decoders for Gaussian parameters and intrinsic attributes, and incorporates environment map encoding to predict direct illumination scaling factors and spherical harmonic coefficients for modeling indirect illumination (Sec.~\ref{sec: pgnr}).
% It leverages geometry-aware decoders and environment map encoding to predict direct illumination scaling factors and spherical harmonic coefficients that model visibility and indirect illumination (Sec.~\ref{sec: pgnr}).
% , avoiding the computational cost of explicit ray tracing (Sec.~\ref{sec: pgnr}).
To ensure strong generalization and realistic rendering, we introduce a 2D-to-3D projection training strategy that reasonably transfers 2D illumination cues into the 3D Gaussian space, simultaneously improving rendering efficiency (Sec.~\ref{sec:training strategy}).

\subsection{Lighting-robust Geometry Refinement}
\label{sec: depth estimation}

To enable a generalizable and relightable 3D Gaussian framework, fast and accurate geometry reconstruction of the target performer is a critical foundation. 
This geometry not only determines the center of each Gaussian point but also serves as an essential component for the subsequent relighting stage. 
However, existing methods~\cite{gao2024relightable, jiang2024gaussianshader, liang2024gs, wu2025deferredgs} typically require per-scene training to optimize Gaussian centers and still struggle to guarantee high-quality and consistent geometry under various lighting.
To address this challenge, we design a Lighting-roubst Geometry Refinement module, consisting of stereo-based depth estimation and depth-guided normal refinement, which jointly produce accurate geometry robust to illumination changes.

\begin{figure}[t]
\vspace{-2mm}
	\centering
	\includegraphics[width=0.90\linewidth]{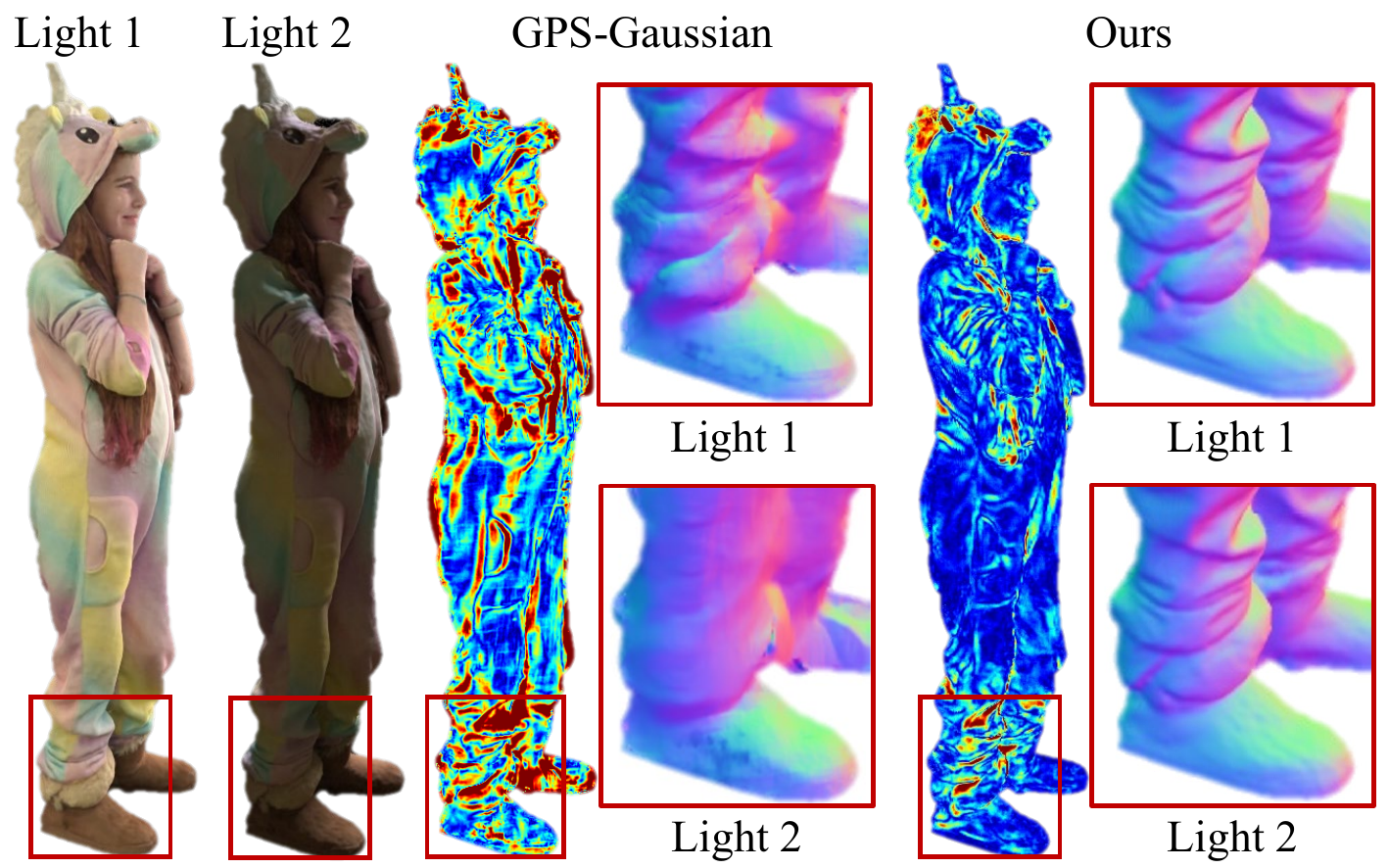}
    % \vspace{-2mm}
    \caption{\textbf{Geometry consistency comparison between GRGS and GPS-Gaussian under varying illumination.} We visualize the normal difference heatmap, where cold colors indicate small angular deviations and hot colors denote large ones. GPS-Gaussian exhibits noticeable degradation under extreme lighting. 
    % For fairness, depth-derived normals are used since GPS-Gaussian does not explicitly estimate surface normals.
    }
	\label{fig:ablation_invariant}
\vspace{-4mm}
\end{figure}

\paragraph{Stereo-based Depth Estimation.}
Inspired by the generalization strategy of GPS-Gaussian~\cite{zheng2024gps}, we leverage RAFT-Stereo~\cite{lipson2021raft}, a stereo-based depth estimator that uses disparity as a geometric constraint across views, enabling consistent depth prediction and improved generalization in diverse human-centric scenarios.
However, pretrained depth estimators tend to be sensitive to illumination variations, leading to unreliable geometry estimation as shown in Fig~\ref{fig:ablation_invariant}.
Therefore, we construct an image encoder trained by a large-scale multi-view relit dataset comprising hundreds of high-quality human scans, enabling the network to learn robust reflectance features for accurate depth estimation across diverse lighting conditions.

Specifically, given $N$ sparse-view images $\{\mathbf{I}_i\}_{i=1}^{N}$ ($\mathbf{I}_i\in\mathbb{R}^{H \times W\times 3}$), captured under an arbitrary lighting of a human-centered scene, we select the two source views nearest to the target camera pose, denoted as $\{\hat{\mathbf{I}}, \bar{\mathbf{I}}\}$, as input for stereo rectification~\cite{papadimitriou1996epipolar}. 
The rectified image pair is then passed through a shared lighting-robust feature extractor $\xi_{\text{img}}$, which is trained on the relit dataset encompassing diverse and varying illumination conditions to explicitly learn features invariant to lighting changes. This yields multi-scale radiance features $\{\hat{\mathbf{F}}_s, \bar{\mathbf{F}}_s\}_{s=1}^K$, where $\hat{\mathbf{F}}_s, \bar{\mathbf{F}}_s \in \mathbb{R}^{\frac{H}{2^s} \times \frac{W}{2^s} \times C}$ denote the feature representations at the $s$-th scale corresponding to the two input images $\{\hat{\mathbf{I}}, \bar{\mathbf{I}}\}$, respectively, and $K$ is the total number of scales.

% The rectified image pair is then passed through a shared lighting-robust feature extractor $\xi_{\text{img}}$ to obtain multi-scale radiance features $\{\hat{\mathbf{F}}_s, \bar{\mathbf{F}}_s\}_{s=1}^K$, where $\hat{\mathbf{F}}_s, \bar{\mathbf{F}}_s \in \mathbb{R}^{\frac{H}{2^s} \times \frac{W}{2^s} \times C}$ denote the feature representations at the $s$-th scale corresponding to the two input images $\{\hat{\mathbf{I}}, \bar{\mathbf{I}}\}$, respectively, $K$ is the total number of scales. 
%
The radiance features mitigate lighting-induced feature mismatches to facilitate accurate depth estimation, as demonstrated in Sec.~\ref{sec:ablation}, while guiding the inference of geometry, material, and light transport within the 3D Gaussian representation (Sec.~\ref{sec: pgnr}).

To balance memory efficiency in 3D correlation construction with rich semantic representation, we feed the final-scale image features  $\{\hat{\mathbf{F}}_K, \bar{\mathbf{F}}_K\}$ and their corresponding camera parameters $\{\hat{\mathbf{G}}, \bar{\mathbf{G}}\}$ into a depth estimation module $\mathcal{G}$ to predict full-resolution depth maps  $\hat{\mathbf{D}}, \bar{\mathbf{D}} \in \mathbb{R}^{H \times W}$ for the selected source view images $\hat{\mathbf{I}}$ and $\bar{\mathbf{I}}$:
\begin{equation}
    \langle\hat{\mathbf{D}}, \bar{\mathbf{D}}\rangle = \mathcal{G}\left(\hat{\mathbf{F}}_K, \bar{\mathbf{F}}_K, \hat{\mathbf{G}}, \bar{\mathbf{G}}\right)
\end{equation}
Within $\mathcal{G}$, a low-resolution 3D correlation volume $\mathbf{\mathcal{M}} \in \mathbb{R}^{\frac{H}{2^K} \times \frac{W}{2^K} \times \frac{W}{2^K}}$ is constructed:
\begin{equation}
    \mathbf{\mathcal{M}}_{ijk} = \sum_{l=1}^C (\hat{\mathbf{F}}_K)_{ijl} \cdot (\bar{\mathbf{F}}_K)_{ikl}
\end{equation}
A GRU-based module is then employed iteratively to predict and refine down-sampled depth maps by querying the correlation volume $\mathbf{\mathcal{M}}$.
Finally, full-resolution, pixel-aligned depth maps are recovered by applying convex upsampling to the refined low-resolution predictions, ensuring spatial consistency and boundary preservation.

\paragraph{Depth-aware Normal Refinement.}
Although the depth map enables 3D point cloud reconstruction via unprojection, they do not capture surface normals, which are critical for accurate shading computations in Sec.~\ref{sec: pgnr}.
To fully harness the depth for surface orientation estimation, we first compute coarse surface normals reconstructed from spatial gradients of a single-view depth map $\mathbf{D} \in \mathbb{R}^{H \times W}$.
% 
% Since normals are computed independently for each view, we simplify the formulation by considering a single-view depth map $\mathbf{D} \in \mathbb{R}^{H \times W}$.
%
Given a foreground pixel $(u, v)$ of $\mathbf{D}$, its corresponding 3D point $\mathbf{X}(u, v) \in \mathbb{R}^3$ is obtained by unprojecting the depth value using the camera projection matrix $\mathbf{P} \in \mathbb{R}^{3 \times 4}$: 
\begin{equation}
    \label{eq: projection}
    \mathbf{X}(u, v) = \Pi_\mathbf{P} \left(u, v, \mathbf{D}\left(u, v\right) \right)
\end{equation}
where $\Pi_\mathbf{P}$ is an unprojection operator defined by $\mathbf{P}$. 
Next, a coarse normal map $\mathbf{N}_c$ is computed via the normalized cross product of horizontal and vertical spatial gradients of the 3D position map $\mathbf{X}$:
% \begin{equation}
%     \mathbf{N}_c(u, v) = \frac{\left(\mathbf{X}(u+1, v) - \mathbf{X}(u, v)\right) \times \left( \mathbf{X}(u, v+1)-\mathbf{X}(u, v)\right)}{\parallel \left(\mathbf{X}(u+1, v) - \mathbf{X}(u, v)\right) \times \left(\mathbf{X}(u, v+1)-\mathbf{X}(u, v)\right) \parallel}
% \end{equation}
\begin{equation}
\resizebox{0.9\linewidth}{!}{$
\mathbf{N}_c(u, v) =
\frac{
(\mathbf{X}(u\!+\!1, v) - \mathbf{X}(u, v))
\times
(\mathbf{X}(u, v\!+\!1) - \mathbf{X}(u, v))
}{
\|
(\mathbf{X}(u\!+\!1, v) - \mathbf{X}(u, v))
\times
(\mathbf{X}(u, v\!+\!1) - \mathbf{X}(u, v))
\|
}
$}
\end{equation}
where $\times$ denotes the cross product.
Due to the convex upsampling, the resulting depth maps lack high-frequency geometric details and exhibit checkerboard artifacts, as illustrated in Sec.~\ref{sec:ablation}.
To address this, we introduce a normal refinement module $\mathcal{N}$ based on a lightweight U-Net architecture, which leverages multi-scale radiance features $\{\mathbf{F}_s\}_{s=1}^K$ and coarse normals $\mathbf{N}_c$ to predict normal offsets $\Delta\mathbf{N} \in \mathbb{R}^3$ and recover fine-grained surface details.
The refinement process can be defined as
\begin{equation}
    \Delta\mathbf{N} = \mathcal{N}\left(\mathbf{N}_c, \{\mathbf{F}_s\}_{s=1}^K\right)
\end{equation}
The refined normal map $\mathbf{N}_f$ is computed by normalizing the sum of $\mathbf{N}_c$ and $\Delta\mathbf{N}$:
\begin{equation}
    \mathbf{N}_f = \frac{\mathbf{N}_c + \Delta\mathbf{N}}{\parallel\mathbf{N}_c + \Delta\mathbf{N}\parallel}
\end{equation}
We denote the normal maps of the two selected source views as $\hat{\mathbf{N}}_f$ and $\bar{\mathbf{N}}_f$.

\subsection{Physically Grounded Neural Rendering}
\label{sec: pgnr}
Based on the accurate geometry, we further propose a PGNR module to combine the generalization capabilities of neural networks with the physical accuracy of physically based rendering, enabling efficient inference of material properties and complex illumination of unseen data. 
\paragraph{Geometry-aware Gaussian Parameter Regression.} 
To support relightable rendering, PGNR parameterizes each 3D Gaussian point $\mathcal{P}$ as a set of attributes:
(1) vanilla attributes: position $\mathbf{p} \in \mathbb{R}^3$, rotation $\mathbf{r} \in \mathbb{R}^4$, scale $\mathbf{s} \in \mathbb{R}^3$, and opacity $\alpha \in \mathbb{R}$,  
(2) geometry-related attributes: surface normal $\mathbf{n} \in \mathbb{R}^3$ and light visibility $\mathbf{v} \in \mathbb{R}^{16}$ encoded via SH coefficients, 
(3) material attributes: albedo $\mathbf{a} \in \mathbb{R}^3$ and roughness $\gamma \in \mathbb{R}$, 
and (4) illumination attributes: indirect lighting $\mathbf{l}_{\text{ind}} \in \mathbb{R}^{48}$ represented by SH coefficients.
We formulate these 3D attributes on corresponding 2D maps via pixel-aligned depth maps (Sec.~\ref{sec: depth estimation}), enabling direct inference of Gaussian point clouds without optimization.
%allowing direct supervision in image space.  
For simplicity, we consider a single-view depth map $\mathbf{D}$ and compute 3D position map $\mathbf{X}$ via unprojection (Eq.~\ref{eq: projection}), while normals $\mathbf{N}_f$ are predicted by the refinement module.
We introduce a geometry-aware encoder $\xi_{\text{geo}}$ that extracts multi-scale geometric features $\{\mathbf{F}_s^g\}_{s=1}^K$ from $\mathbf{D}$ and $\mathbf{N}_f$, encoding both 2D image features and 3D spatial geometry. 
These features are then fused with radiance features to incorporate appearance and geometric context for jointly modeling geometry-aware attribute maps.
Specifically, we first extract the full-resolution Gaussian features $\Theta$ through a decoder $\Phi_{\text{parm}}$:
\begin{equation}
    \Theta=\Phi_{\text{parm}}\left(\{\mathbf{F}_s\}_{s=1}^K \oplus \{\mathbf{F}_s^g\}_{s=1}^K\right)
\end{equation}
where $\oplus$ denotes the concatenation operation applied across all feature levels.
Subsequently, the vanilla Gaussian parameter maps (rotation map $\mathbf{R}$, scale map $\mathbf{S}$, and opacity map $\mathbf{O}$) are predicted with a Gaussian head $h_g$:
\begin{equation}
    \langle\mathbf{R}, \mathbf{S}, \mathbf{O}\rangle = h_g\left(\Theta\right)
\end{equation}
% As for material attributes, to accelerate the convergence of albedo estimation under diverse lighting conditions, we introduce a residual component $\Delta\mathbf{A}$, which serves as a delighting term to better disentangle shadows and illumination effects from intrinsic surface properties. 
For material attributes, we introduce a residual component $\Delta\mathbf{A}$ to accelerate the convergence of albedo estimation under diverse lighting conditions. This component acts as a delighting term, helping disentangle shadows and illumination effects from the intrinsic surface properties.
We achieve this by leveraging a material prediction head $h_m$ to jointly estimate the residual albedo $\Delta\mathbf{A}$ and the surface roughness map $\mathbf{\Gamma}$:
\begin{equation}
    \langle\Delta\mathbf{A}, \mathbf{\Gamma}\rangle = h_m\left(\Theta\right)
\end{equation}
The final albedo is computed as: $\mathbf{A} = \text{Sigmoid}\left(\mathbf{I} + \Delta\mathbf{A}\right)$, where $\mathbf{I}$ denotes the input image.
Besides, since the SH-encoded visibility map $\mathbf{Y}_{\text{vis}}$ is independent of image appearance, we predict it via another lightweight decoder $\Phi_{\text{vis}}$ conditioned solely on the geometric features $\{\mathbf{F}_s^g\}_{s=1}^K$.
\begin{figure*}
\vspace{-15pt}
	\centering
	\includegraphics[width=\textwidth]{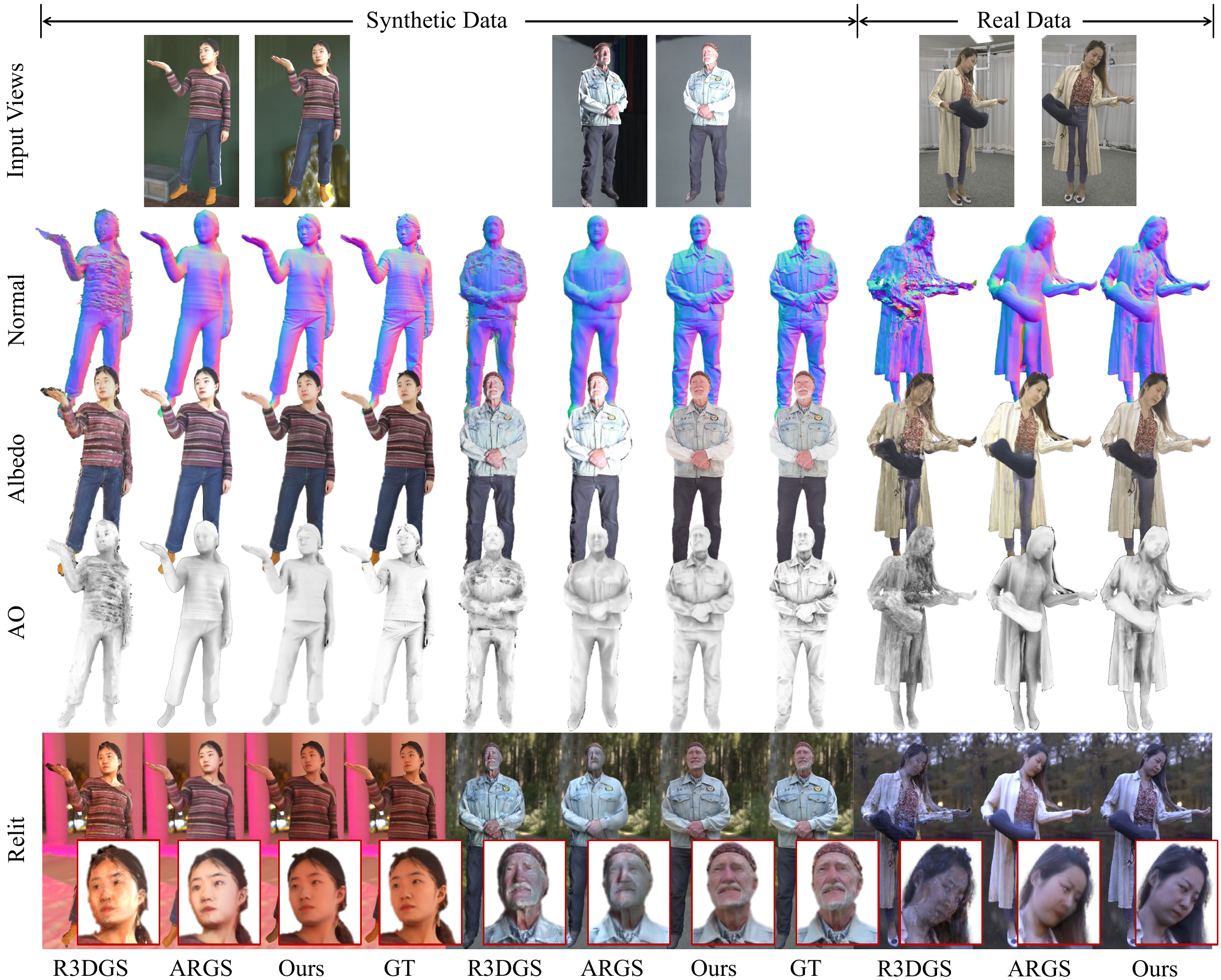}
	% \vspace{-16pt}
	\caption{Qualitative comparison of our method and 3DGS-based methods. Zoom in for the best view.}
	\label{fig:camparison_3d}
% \vspace{-5pt}
\end{figure*}
\paragraph{Light Parameterization.}
\label{sec: light parametrization}
Accurate shading in 3D space typically requires dense sampling of incoming light directions, leading to substantial computational and memory overhead. 
To mitigate this, we prefilter the high-resolution HDR environment map $\mathbf{L}_{\text{d}}\in \mathbb{R}^{H_{\text{h}} \times W_{\text{h}} \times 3}$ to obtain a convolved version $\mathbf{L}_{\text{d}}^\prime \in \mathbb{R}^{H_{\text{l}} \times W_{\text{l}} \times 3}$ that approximates the integral of incident illumination:
\begin{equation}
    \mathbf{L}_{\text{d}}^{\prime}\left(\boldsymbol{\omega}^{\prime}\right) = \int_{\Omega} \left(\boldsymbol{\omega}^{\prime} \cdot \boldsymbol{\omega}\right) ^l\mathbf{L}_{\text{d}}\left(\boldsymbol{\omega}\right)d\boldsymbol{\omega}
\end{equation}
where $\boldsymbol{\omega}$ and $\boldsymbol{\omega}^{\prime}$ denote spherical directions in the original and convolved environment map respectively, $\Omega$ represents the unit sphere, and $l$ is the shininess exponent controlling the angular falloff in the Phong reflectance model. 
However, such integration may result in underexposed lighting compared to ground-truth shading, as illustrated in Sec.~\ref{sec:ablation}. 
To address this, we introduce a direct illumination scaling factor $s_\text{d} \in \mathbb{R}$ to globally compensate for the brightness discrepancy.
In addition, since the SH-encoded indirect lighting map $\mathbf{Y}_{\text{ind}}$ is influenced not only by the appearance and geometry but also by the input illumination, we design a light encoder-decoder that predicts the direct illumination scaling map $\mathbf{S}_\text{d}$  and the indirect lighting map $\mathbf{Y}_{\text{ind}}$:
\begin{equation}
       \langle\mathbf{S}_\text{d}, \mathbf{Y}_{\text{ind}}\rangle = h_{l}\left(\Phi_{l}\left(\xi_{l}\left(\mathbf{L}_{\text{d}}\right)\oplus \mathbf{L}_{\text{d}}^{\prime} \right) \oplus \Theta \right)
\end{equation}
where $\xi_l$, $\Phi_l$, and $h_l$ denote the light encoder, decoder, and output head, respectively. Then, $s_\text{d}$ is obtained by spatially averaging $\mathbf{S}_\text{d}$ over the foreground $N_f$:
\begin{equation}
    s_\text{d} = \frac{\sum_{i=1}^{N_f}\mathbf{S}_\text{d}\left(u_i, v_i\right) }{N_f}
\end{equation}

\paragraph{Physically-based Rendering.}
To enable physically plausible light interaction on the human surface, we compute the PBR color $\mathbf{C}_{\text{PBR}}$ for each Gaussian point. Following the rendering equation~\cite{kajiya1986rendering}, the outgoing radiance $\mathbf{C}_{\text{PBR}}$ in direction $\boldsymbol{\omega}_o$ can be given by:
% \begin{equation}
%     \mathbf{C}_{\text{PBR}}\left(\boldsymbol{\omega}_o\right) = \int_\Omega \mathbf{L}\left(\boldsymbol{\omega}_i\right) f\left(\boldsymbol{\omega}_i, \boldsymbol{\omega}_o, \mathbf{a}, \gamma \right) \left(\boldsymbol{\omega}_i \cdot \boldsymbol{n}\right)d\boldsymbol{\omega}_i
% \end{equation}
\begin{equation}
\mathbf{C}_{\text{PBR}}(\boldsymbol{\omega}_o)
=\!\!\int_\Omega\!\!
\mathbf{L}(\boldsymbol{\omega}_i)
f(\boldsymbol{\omega}_i,\boldsymbol{\omega}_o,\mathbf{a},\gamma)
(\boldsymbol{\omega}_i\!\cdot\!\boldsymbol{n})\,d\boldsymbol{\omega}_i
\end{equation}
where  $f$ is the simplified Disney BRDF function~\cite{burley2012physically} modeling the surface reflectance properties, and $\Omega$ here represents the hemisphere oriented around the surface normal $\boldsymbol{n}$.
$\mathbf{L}\left(\boldsymbol{\omega}_i\right)$ denotes the incident radiance from direction $\boldsymbol{\omega}_i$:
% \begin{equation}
%     \mathbf{L}\left(\boldsymbol{\omega}_i\right) = V\left(\boldsymbol{\omega}_i\right) \left(s_{\text{d}} \cdot \mathbf{L}_{\text{d}}^{\prime}\left(\boldsymbol{\omega}_i\right)\right) + \left(1 - V\left(\boldsymbol{\omega}_i\right)\right)\mathbf{L}_{\text{ind}}\left(\boldsymbol{\omega}_i\right)
% \end{equation}
\begin{equation}
\mathbf{L}(\boldsymbol{\omega}_i)
=\!V(\boldsymbol{\omega}_i)\!\left(s_{\text{d}}\mathbf{L}_{\text{d}}^{\prime}(\boldsymbol{\omega}_i)\right)
+\!\left(1\!-\!V(\boldsymbol{\omega}_i)\right)\!\mathbf{L}_{\text{ind}}(\boldsymbol{\omega}_i)
\end{equation}
where $V\left(\boldsymbol{\omega}_i\right)$ denotes the light visibility from direction $\boldsymbol{\omega}_i$, parameterized using SH coefficients $\mathbf{v}$. Similarly, $\mathbf{L}_{\text{ind}}\left(\boldsymbol{\omega}_i\right)$ represents the indirect illumination from direction $\boldsymbol{\omega}_i$, also encoded via SH coefficients $\mathbf{l}_{\text{ind}}$.
Although R3DGS~\cite{gao2024relightable} employs a similar lighting model, it estimates indirect illumination in an implicit manner through unsupervised inverse rendering, which often results in inaccurate predictions and limited generalization to novel lighting conditions while imposing a substantial computational burden due to the explicit ray tracing strategy. 
In contrast, our GRGS utilizes the strong generalization capability of a U-Net-based architecture in conjunction with a 2D-to-3D projection training strategy, enabling more accurate estimation of both visibility and indirect illumination under arbitrary lighting conditions. 
After computing the physically-based rendering color $\mathbf{C}_{\text{PBR}}$ for each Gaussian point, we perform rasterization to synthesize the final image.

\subsection{2D-to-3D Projection Training}
\label{sec:training strategy}
Leveraging the excellent differentiability of 3DGS~\cite{kerbl20233d}, GRGS optimizes lighting parameters in the 3D space directly from 2D image supervision, thereby circumventing the costly computation of explicit ray tracing.
% GRGS directly projects 2D image supervision into the 3D space to avoid heavy computation of explicit ray tracing.
% 
Specifically, ground truths of ambient occlusion, direct lighting, and indirect lighting maps are used as photometric supervision signals to guide the optimization of the 3D Gaussian representation via efficient gradient-based learning.
Among these signals, ambient occlusion and direct lighting maps are jointly used to provide more accurate supervision for visibility estimation. While ambient occlusion captures geometric occlusion effects, direct lighting offers complementary shadow-related illumination cues, and their combination helps enforce visibility consistency between 2D renderings and the underlying 3D geometry.
% Using only ground-truth ambient occlusion as supervision may lead to visibility estimates that appear correct in 2D rendering but are inconsistent in 3D space. To address this, we incorporate ground-truth direct light as an additional supervision signal, which enhances the alignment between 2D projections and 3D geometry. By providing extra shadow-related illumination cues, direct light supervision helps enforce consistent visibility implicitly.
%
For effective indirect lighting learning within the rendering equation, we employ a gradient-truncated hard shadow fusion scheme, which truncates gradients through visibility to prevent erroneous gradient propagation and leverages hard shadow fusion to accelerate the convergence of indirect lighting.

\section{Experiments}
\paragraph{Implementation Details.}
Our framework is trained on a single NVIDIA RTX 4090 GPU over approximately four days. We first train the LGR module for 100K iterations, where both the human subject and illumination conditions are randomly sampled in each iteration. Next, we train the entire framework for 300K iterations to jointly optimize for high-quality geometry reconstruction and realistic relighting performance. More details of the optimization process are provided in Supp.Mat..

\paragraph{Datasets.}
We utilize two human scan datasets, Twindom~\cite{twindom} and THuman2.0~\cite{yu2021function4d}, to validate the effectiveness of our method. 
To ensure high-quality rendering, we carefully selected 800 scans from Twindom and 337 scans from THuman2.0, filtering out meshes with noticeable artifacts. 
Additionally, 384 HDR environment maps were collected from Polyhaven, HDRMAPS, and iHDRI to simulate diverse illumination conditions. More dataset construction details are listed in Supp.Mat..
In addition, we also employ multi-view real-world data from GPS-Gaussian~\cite{zheng2024gps} and DNA-Rendering~\cite{cheng2023dna} to further evaluate the generalization ability of our method in real scenarios.

\newcommand{\tablecomparison}{
\begin{table*}[t]
% \footnotesize
\scriptsize
\centering
\definecolor{Gray}{gray}{0.85}
\vspace{-5mm}
\caption{ 
\textbf{Quantitative comparison on the synthetic dataset.} In accordance with~\cite{zheng2024gps}, the SSIM and LPIPS metrics are calculated within the bounding box delineating the human region, while the MAE of normal and PSNR metrics is computed within the foreground mask.
}
\vspace{-2mm}
\label{tab:comparison}
\resizebox{\textwidth}{!}{
\begin{tabular}{lcccccccccccc}
\toprule
& \multicolumn{1}{c}{Normal} & \multicolumn{3}{c}{Diffuse Albedo} & \multicolumn{3}{c}{Ambient Occlusion} & \multicolumn{3}{c}{Relighting}\\
& MAE$\downarrow$  & PSNR$\uparrow$  & SSIM$\uparrow$  & LPIPS$\downarrow$  & PSNR$\uparrow$  & SSIM$\uparrow$  & LPIPS$\downarrow$  & PSNR$\uparrow$  & SSIM$\uparrow$  & LPIPS$\downarrow$ \\
\midrule
R3DGS~\cite{gao2024relightable} & 10.208 & 23.584 & 0.829 & 0.165& 20.983 & 0.761 & 0.388& 21.983 & 0.812 & 0.162 \\
ARGS~\cite{li2024animatable}  & 6.941  & 25.937 & 0.865 & 0.142 & 23.721 & 0.837 & 0.129 & 23.879 & 0.846 & 0.147 \\

\midrule
% w/o Pose Opt. & 27.71 & 0.9713 & 0.0218 & 25.09 & 0.9614 & 0.0306 & 24.80 & 0.9609 & 0.0310 & 26.16 & 0.9610 & 0.0227 \\
%
Ours & \textbf{5.369} &  \textbf{27.536} & \textbf{0.936} & \textbf{0.080}  & \textbf{24.470} & \textbf{0.865} & \textbf{0.105} & \textbf{27.977} & \textbf{0.926} & \textbf{0.099}  \\
\bottomrule

\end{tabular}
}
\vspace{-3.5mm}
% }
\end{table*}
% }

}
\tablecomparison
\begin{figure}
% \vspace{-2pt}
	\centering
	\includegraphics[width=\linewidth]{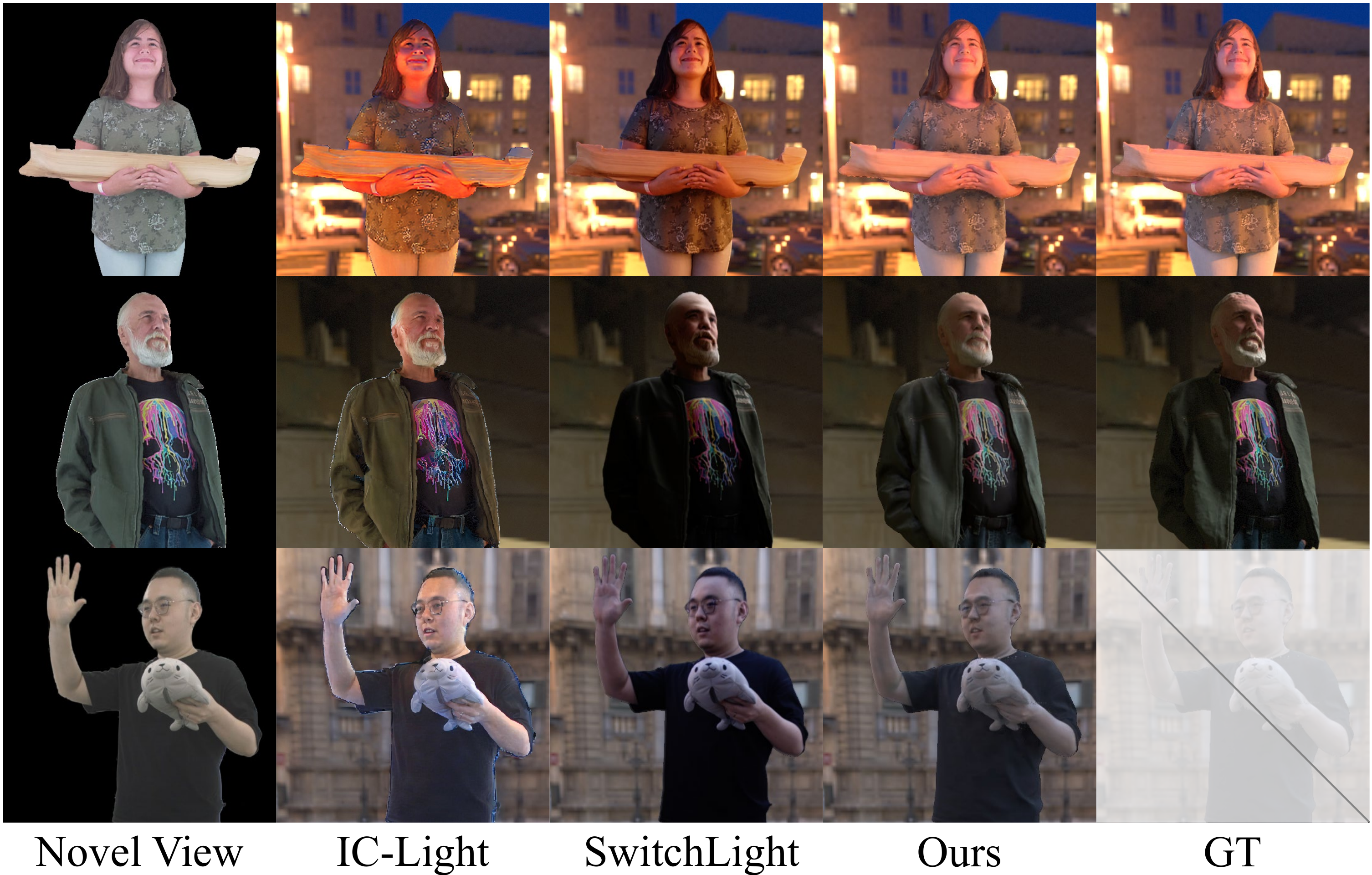}
	\vspace{-15pt}
	\caption{Qualitative comparison between our method and 2D image-based approaches. The bottom row shows results from the real dataset~\cite{zheng2024gps}. Zoom in for the best view.}
	\label{fig:camparison_2d}
 \vspace{-15pt}
\end{figure}

\paragraph{Baselines and Metrics.}
We compare our method on our synthetic dataset and real dataset~\cite{zheng2024gps} against two 3DGS-based approaches, R3DGS~\cite{gao2024relightable} and ARGS~\cite{li2024animatable}, as well as two 2D image-based methods, IC-Light~\cite{zhang2025scaling} and SwitchLight~\cite{kim2024switchlight}. 
Note that since 2D methods cannot synthesize novel views, we directly use ground-truth novel view images as their input for a fair comparison.
For quantitative evaluation, we employ PSNR, SSIM, and LPIPS to evaluate the diffuse albedo, ambient occlusion, and relighting quality. Further, we utilize MAE to evaluate the normal map.

\subsection{Comparison Results}
% \begin{figure}
% % \vspace{-6mm}
% 	\centering
% 	\includegraphics[width=0.8\linewidth]{figs/ablation_invariant.pdf}
%     % \vspace{-5mm}
%     \caption{Geometry consistency comparison.}
% 	\label{fig:ablation_invariant}
% % \vspace{-6mm}
% \end{figure}
We first make a comparison with 3D-based methods (\textit{i.e.}, R3DGS~\cite{gao2024relightable} and ARGS~\cite{li2024animatable}) from the perspective of material estimation, geometry reconstruction, and relighting quality, as shown in Fig.~\ref{fig:camparison_3d} and Table~\ref{tab:comparison}.
%
% R3DGS is a general-purpose inverse rendering approach designed for arbitrary scenes. It struggles to recover smooth and detailed surface normals from sparse-view inputs due to the absence of human-specific geometric priors. Consequently, the reconstructed surfaces are overly coarse, resulting in suboptimal relighting performance under varying illumination. 
R3DGS~\cite{gao2024relightable} struggles to recover smooth, detailed normals from sparse views without human-specific priors, leading to suboptimal relighting performance.
ARGS~\cite{li2024animatable} alleviates this by introducing a body template prior. However, optimizing geometry solely from rendered images leads to overly smooth surfaces lacking high-frequency details.
Moreover, both methods depend on inverse rendering under an inherently ill-posed formulation, which yields inaccurate material estimation and consequently degrades the overall quality of relighting results.
Thanks to the proposed LGR and PGNR modules, GRGS produces high-quality geometry and realistic relighting results under novel viewpoints. 

Furthermore, a comparison with representative 2D-based relighting methods (\textit{i.e.}, IC-Light~\cite{zhang2025scaling} and SwitchLight~\cite{kim2024switchlight}) is illustrated in Fig.~\ref{fig:camparison_2d}. 
The baseline methods produce noticeable shading and lighting variations that often lack physical plausibility, exhibiting artifacts such as inconsistent light–geometry interactions and unnatural shadow orientations, thereby diminishing overall realism. 
In contrast, our results closely match the ground-truth images generated by a physically-based renderer and deliver superior perceptual quality. Additional evidence is provided by a user study (see Supp.Mat.).
%In contrast, our results closely match the ground-truth images generated by a physically-based renderer and deliver superior perceptual quality, as further evidenced by a user study (see Supp.Mat.). 
%in which participants consistently rated our relighting outputs as more visually convincing and physically coherent than those of baseline methods.
% Further, the comparison with 2D-based methods (\textit{i.e.}, IC-Light~\cite{zhang2025scaling} and SwitchLight~\cite{kim2024switchlight}) are illustrated in Fig.~\ref{fig:camparison_2d}.
% The baseline methods produce noticeable shading and lighting variations, which, however, lack physical plausibility, thereby undermining overall realism.
% In contrast, our results closely approximate the ground-truth images generated by a physically-based renderer and achieve superior perceptual quality. 
% This is further corroborated by a user study (see Supp.Mat.), which confirms that our approach yields more visually plausible and realistic relighting results than baseline methods.

% \vspace{-4mm}
% \label{sec: ablation}
\begin{figure}[!htpb]
\vspace{-6pt}
	\centering
	\includegraphics[width=0.9\linewidth]{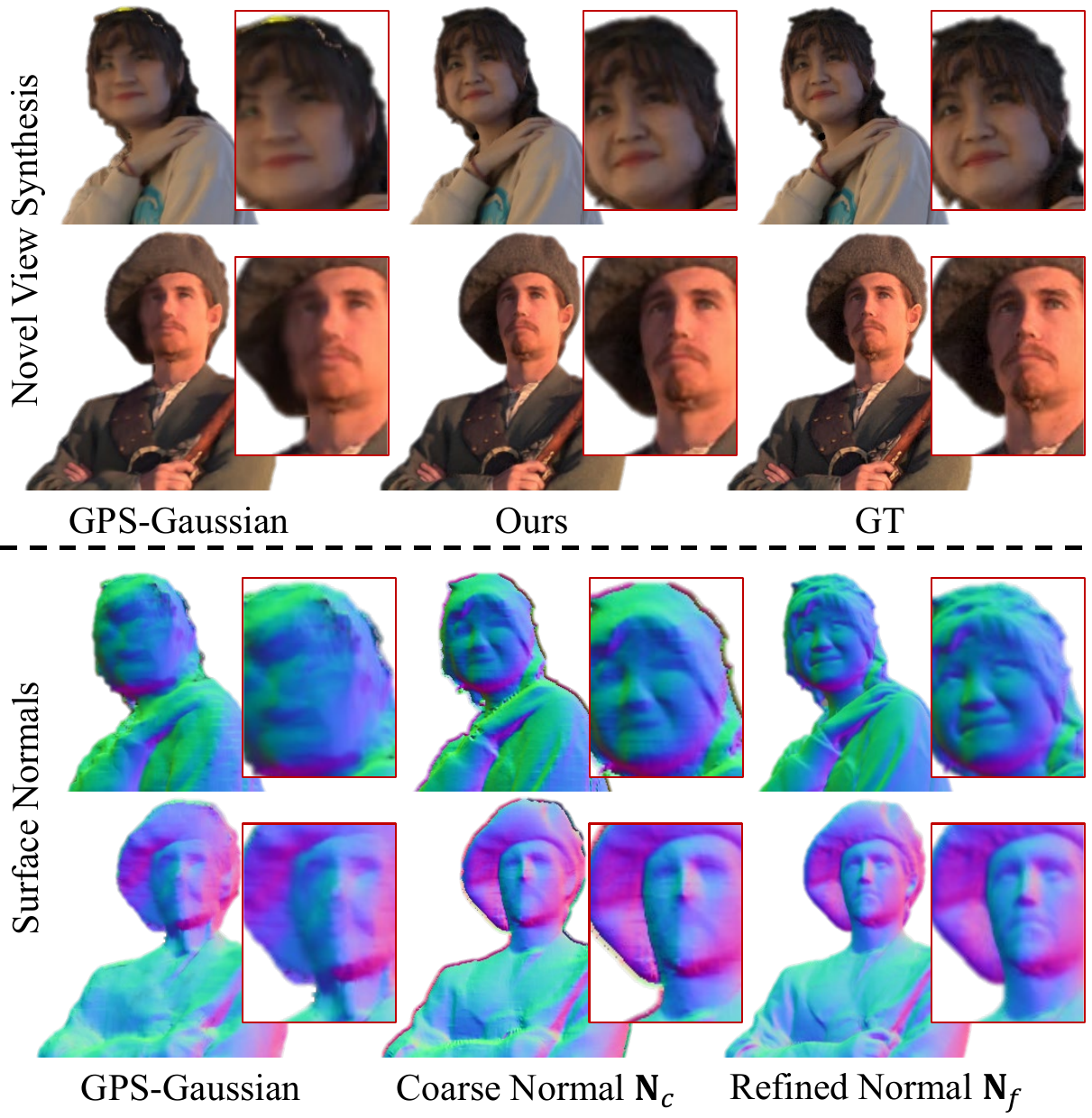}
	\vspace{-7pt}
	\caption{Enhanced novel view rendering and surface normals results via LGR. Zoom in for the best view.}
	\label{fig:ablation_geo}
 \vspace{-15pt}
\end{figure}

% \input{tables/nvs_depth}
% \nvsdepth
\subsection{Ablation Study}
\label{sec:ablation}
\paragraph{Geometry reconstruction.}
% \paragraph{Lighting-invariant Prior.}
A major limitation of the original GPS-Gaussian~\cite{zheng2024gps} is that it struggles to maintain robustness against uneven lighting on the human surface without lighting priors, leading to a significant degradation in geometric consistency, as shown in  Fig.~\ref{fig:ablation_invariant}.
In contrast, our LGR module learns lighting-robust radiance features under diverse illumination conditions, thereby improving reconstruction accuracy under uneven lighting and enhancing both geometric consistency and NVS quality (Table.~\ref{table:nvs}).

% \paragraph{Normal Refinement.}
Although the lighting-robust radiance features enhance depth estimation accuracy, the use of convex upsampling for full-resolution recovery inevitably causes the loss of high-frequency geometric details and introduces checkerboard artifacts in the coarse normal map $\mathbf{N}_c$, as shown in Fig.~\ref{fig:ablation_geo}.
The proposed normal refinement module leverages both semantic information from radiance features and geometric cues from the coarse normal map to reconstruct high-quality surface normals with consistent improvements in qualitative (Fig.~\ref{fig:ablation_geo}) and quantitative (see Supp.Mat.) results.

\newcommand{\nvsdepth}{
\noindent
\begin{table}[!ht]%[b]%[!htbp]
    % \centering
    \vspace{-7pt}
    \caption{NVS and Depth Evaluation.}
    \label{tab:lighting_prior}
    \vspace{-7pt}
    \small
    \resizebox{\linewidth}{0.98cm}{  % 设置相同高度
    \begin{tabular}{l|ccc|cc}
    \toprule
    \multirow{2}{*}{Method} & \multicolumn{3}{c|}{NVS} & \multicolumn{2}{c}{Depth} \\
    & PSNR$\uparrow$ & SSIM$\uparrow$ & LPIPS$\downarrow$ & EPE$\downarrow$ & 1 pix$\uparrow$ \\
    \midrule
    GPS-Gaussian~\cite{zheng2024gps} & 30.444 & 0.913 & 0.107  & 1.463 & 68.91\\
    Ours   & \textbf{31.694} & \textbf{0.949} & \textbf{0.067}  & \textbf{0.692} & \textbf{85.39}\\
    \bottomrule
    \end{tabular}
    }
    \label{table:nvs}
\end{table}
% \hfill
% \begin{minipage}[t]{0.33\textwidth}
%     \centering
%     \captionof{table}{MAE Comparison}
%     \vspace{-7pt}
%     \resizebox{\textwidth}{0.98cm}{  % 设置高度为 4.5cm
%     \begin{tabular}{l|c}
%     \toprule  
%     Method & MAE $\downarrow$\\ 
%     \midrule
%     GPS-Gaussian~\cite{zheng2024gps} & 8.145    \\ 
%     Coarse Normal $\mathbf{N}_c$ &  5.903   \\ 
%     Refined Normal $\mathbf{N}_f$ & \textbf{5.369}  \\
%     \bottomrule
%     \end{tabular}
%     }
%     \label{table:mae}
% \end{minipage}

}

\nvsdepth
\begin{figure}[!ht]
\vspace{-2mm}
	\centering
	\includegraphics[width=\linewidth]{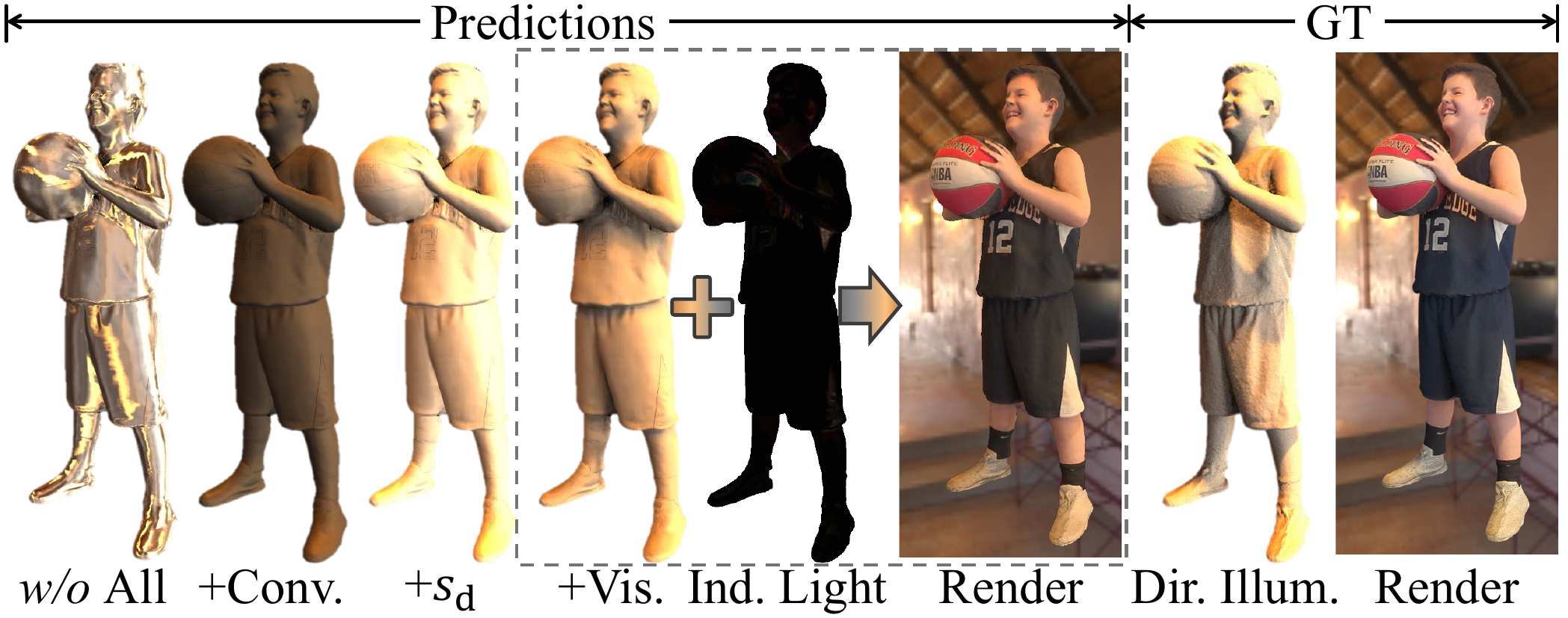}
	\vspace{-2mm}
	\caption{Enhanced light transport through PGNR. Zoom in for the best view.}
	\label{fig:ablation_light}
 \vspace{-4mm}
\end{figure}
\vspace{-6mm}
\paragraph{Light Transport.}
To mitigate the brightness loss caused by convolved environment lighting, we introduce a feed-forward direct illumination scaling factor $s_{\text{d}}$, which yields shading closer to the ground truth as shown in Fig.~\ref{fig:ablation_light}.
% We introduce a feed-forward direct illumination scaling factor $s_{\text{d}}$ to compensate for the brightness loss from convolved environment lighting. As shown in Fig.~\ref{fig:ablation_light}, applying $s_{\text{d}}$ brings the shading closer to the ground truth. 
%
To further enhance realism by modeling shadow effects and complex light interactions such as multi-bounce illumination, we additionally learn light visibility and indirect lighting. As illustrated in the figure, 
our method accurately captures occlusion shadows from the arm and ball.
% our method accurately captures occlusion-induced shadows, including those cast by the arm and ball.
%
Across all experiments, realistic effects are achieved with only tens of light samples per Gaussian. Specifically, sampling 40 rays strikes an effective balance between quality and efficiency. With TensorRT acceleration, our model runs at 20 FPS during inference.
Further analysis of computational efficiency is provided in Supp.Mat..

\section{Conclusion}
This paper proposes GRGS, a generalizable and relightable 3D Gaussian framework for high-fidelity human novel view synthesis under diverse lighting conditions.
GRGS adopts a supervised 2D-to-3D projection strategy to transfer geometry, material, and illumination cues from multi-view 2D observations into 3D Gaussian attributes, enabling efficient feed-forward inference without person-specific optimization.
To ensure accurate geometry, we first construct a Lighting-robust Geometry Refinement module to estimate robust depth and surface normals.
In addition, a Physically Grounded Neural Rendering module is presented to integrate physics-based shading with neural inference to synthesize realistic lighting effects, avoiding the cost of explicit ray tracing.
Thanks to these designs, GRGS achieves high-quality, editable human relighting and strong generalization to unseen identities and lighting environments.
{
    \small
    \bibliographystyle{ieeenat_fullname}
    \bibliography{main}
}

\clearpage
\setcounter{page}{1}
\maketitlesupplementary
In this supplementary material, we provide additional information, including preliminary concepts (Sec.\ref{sec:preliminary}),  implementation details (Sec.\ref{sec:implementation}), more comparison results  (Sec.\ref{sec:More Comparison Results}), more analysis on core component (Sec.\ref{sec:extra experiment}), detailed loss functions(Sec.\ref{sec:optimization}) and a discussion of limitations (Sec.~\ref{sec:limitations}).

\section{Preliminary}
\label{sec:preliminary}
\paragraph{3D Gaussian Splatting.} 3DGS~\cite{kerbl20233d} is an explicit point-based representation that models a scene as a set of 3D Gaussians. Each Gaussian is parameterized by its position $\mathbf{p}$, a covariance matrix $\mathbf{\Sigma}$ (constructed from a rotation vector $\mathbf{r}$ and a scale vector $\mathbf{s}$), opacity $\alpha$, and color $\mathbf{c}$. In our framework, the color $\mathbf{c}$ can be formulated to represent any attribute requiring optimization under 2D-to-3D supervision. Then each gaussian can be expressed as:
\begin{equation}
f(\mathbf{x}|\mathbf{p},\mathbf{\Sigma})=\exp\left(-\frac{1}{2} (\mathbf{x} - \mathbf{p})^\top \mathbf{\Sigma}^{-1} (\mathbf{x} - \mathbf{p})\right)
    \label{eq: gaussian pdf}
\end{equation}
where the constant factor in Eq.~\ref{eq: gaussian pdf} is omitted. A 2D image is rendered through rasetrization. Specifically, the 3D Gaussians are projected onto 2D planes, resulting in 2D Gaussians. The pixel color $\mathbf{C}$ is determined by blending $N$ ordered 2D Gaussians that overlap this pixel:
\begin{equation}
    \mathbf{C} = \sum_{i=1}^N \alpha_i \prod_{j=1}^{i-1}(1-\alpha_j)\mathbf{c}_i
    \label{eq: rasterization}
\end{equation}
where $\mathbf{c}_i$ is the color of each 2D Gaussian, and $\alpha_i$ is the blending weight derived from the learned opacity and 2D Gaussian distribution \cite{zwicker2001surface}.
\paragraph{BRDF modelling.}
The whole BRDF~\cite{burley2012physically} employed in our method is composed of a diffuse term $f_\text{d}$ and a specular term $f_{\text{s}}$, defined as:
% \begin{equation}
%     f\left(\boldsymbol{\omega}_i, \boldsymbol{\omega}_o, \mathbf{a}, \gamma \right) = \underbrace{\frac{\mathbf{a}}{\pi}}_{f_d} + \underbrace{\frac{D\left(\mathbf{h}; \gamma \right) \cdot F\left(\boldsymbol{\omega}_o, \mathbf{h}\right)\cdot G\left(\boldsymbol{\omega}_i ,\boldsymbol{\omega}_o, \mathbf{h}; \gamma \right)}{ \left(\boldsymbol{\omega}_i \cdot \boldsymbol{n}\right)\cdot \left(\boldsymbol{\omega}_o \cdot \boldsymbol{n}\right)}}_{f_s}
% \end{equation}
\begin{equation}
f(\boldsymbol{\omega}_i,\boldsymbol{\omega}_o,\mathbf{a},\gamma)
=
\underbrace{\tfrac{\mathbf{a}}{\pi}}_{f_d}
+
\underbrace{
\tfrac{
D(\mathbf{h};\gamma)
F(\boldsymbol{\omega}_o,\mathbf{h})
G(\boldsymbol{\omega}_i,\boldsymbol{\omega}_o,\mathbf{h};\gamma)
}{
(\boldsymbol{\omega}_i\!\cdot\!\boldsymbol{n})
(\boldsymbol{\omega}_o\!\cdot\!\boldsymbol{n})
}
}_{f_s}
\end{equation}
where $\mathbf{h} = \frac{\left(\boldsymbol{\omega}_i + \boldsymbol{\omega}_o\right)}{2}$ denotes the half-vector between the incoming direction $\boldsymbol{\omega}_i$ and outgoing direction $\boldsymbol{\omega}_o$. The functions $D\left(\cdot\right)$, $F\left(\cdot\right)$, and $G\left(\cdot\right)$ correspond to the normal distribution function (NDF), Fresnel term, and geometric term, respectively. The parameter $\mathbf{a}$ represents the albedo, while $\gamma$ denotes the roughness.

\begin{figure*}[t]
% \vspace{-10pt}
	\centering
	\includegraphics[width=\textwidth]{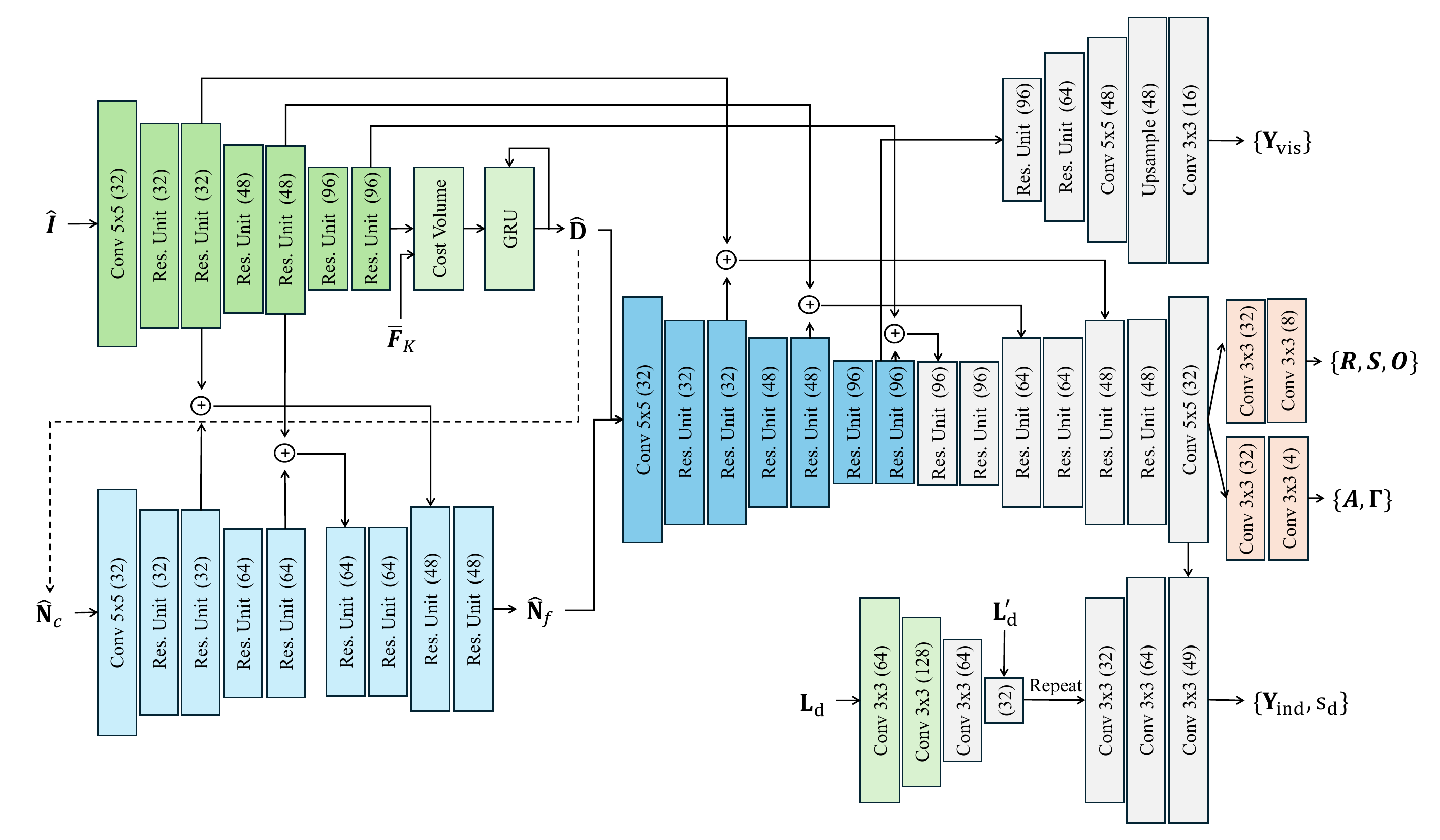}
	% \vspace{-15pt}
	\caption{\textbf{Network architecture.} For simplicity, we consider a single-view image $\hat{I}$ as the input, while $\bar{F}_K$ denotes the final-scale image feature extracted from the other view.}
	\label{fig:network}
 % \vspace{-15pt}
\end{figure*}

\section{Implementation Details}
\label{sec:implementation}

\paragraph{Dataset construction.}
We construct our dataset following a similar multi-view setup as GPS-Gaussian~\cite{zheng2024gps}, but unlike GPS-Gaussian, which only renders novel view RGB images, our data construction explicitly accounts for geometry, material, and illumination information.
We employ Cycles renderer in Blender to render the datasets, positioning 16 cameras uniformly in a circular arrangement with an angular interval of 22.5\degree \ between adjacent cameras.
Each human scan is rendered from every input camera pair, along with three novel views sampled from the intersection arc between the input cameras, under five randomly selected HDR environment maps. 
For each rendering, we generate corresponding outputs, including albedo, normal, depth, foreground mask, ambient occlusion, indirect light, shading, and relighting images.
Since the scans are captured under uniformly illuminated conditions, their texture maps are approximately treated as intrinsic albedo in this dataset.
Our training set consists of 600 scans from the Twindom~\cite{twindom} dataset and 277 scans from the THuman2.0~\cite{yu2021function4d} dataset. For evaluation, we reserve 200 scans from Twindom and 60 scans from THuman2.0 as the test set.

\paragraph{Network Architecture and Hyperparameter.}
The detailed network architecture is illustrated in Fig.~\ref{fig:network}. In the LGR module, we set the number of radiance feature scales $K=3$, with feature dimensions of 32, 48, and 96, respectively, for$\{\mathbf{F}_s\}_{s=1}^K$. 
In the PGNR module, the target environment map is convolved from a resolution of $1024\times 512$ down to $32\times 16$ using a shininess exponent of $l=16$. 
Owing to the efficient light parameterization, we sample only 40 rays per Gaussian point oriented around the surface normal $\boldsymbol{n}$ for PBR, enabling fast and high-quality shading under arbitrary environment maps.
We train our GRGS framework using the AdamW~\cite{loshchilov2017decoupled} optimizer with an initial learning rate of $2e^{-4}$. For the LGR module, we use a batch size of 2, while a batch size of 1 is adopted for the PGNR module.
Note that during PGNR training, the LGR module is jointly optimized to facilitate more accurate localization of Gaussian point positions.

\newcommand{\efficiencyablation}{
\begin{table}[t]
    \centering
    \caption{Runtime \& relighting comparisons between spherical harmonics-based visibility (SH-Vis) modeling and ray-traced visibility (RT-Vis) under various ray sampling configurations.}
    \small
    \setlength\tabcolsep{2pt} 
    \begin{tabular}{l|cccccc}
    \toprule  
     & \multicolumn{3}{c}{SH-Vis} & \multicolumn{3}{c}{RT-Vis} \\
    Sampling No. & 40 & 128 & 256 & 40 & 128 & 256 \\
    \midrule
    Runtime $\downarrow$ & \textbf{49} & 71 & 103 & 230 & 819 & 1536\\
    \midrule
    Relit PSNR $\uparrow$ & 27.977 & 27.968 & \textbf{28.041} & 26.762 & 27.593 & 27.812\\
    \bottomrule
    \end{tabular}
    \label{table:efficiency_ablation}
\end{table}
}
\efficiencyablation

\section{More Comparison Results}
\label{sec:More Comparison Results}
\paragraph{Computation efficiency.}
To quantitatively validate the computational efficiency of our method, we conduct runtime and relighting comparisons between our spherical harmonics-based visibility (SH-Vis) modeling and ray-traced visibility (RT-Vis) proposed in R3DGS under various ray sampling configurations (40, 128, 256) as shown in Table~\ref{table:efficiency_ablation}.
The results demonstrate that our SH-Vis design achieves a significant speedup (at least 4× faster), compared to ray-traced visibility, while also delivering better relighting quality. Although SH-Vis-256 yields slightly higher image quality, its runtime doubles compared to SH-Vis-40, which hinders real-time inference. Moreover, SH-Vis-40 achieves comparable PSNR due to effective light parametrization, making it a more practical choice for our experiments.
In contrast, RT-Vis incurs substantially longer runtimes and yields lower relighting quality, primarily because Gaussian point clouds lack explicit surfaces, forcing ray tracing to rely on computationally expensive volumetric intersection, which is not yet as accurate or efficient as mesh-based methods. In comparison, SH-Vis can learn high-frequency details from ground truth supervision, enabling it to achieve more accurate and visually appealing relighting results.

\newcommand{\efficiencycomparison}{
\begin{table}[t]
    \centering
    \caption{Runtime \& relighting comparisons against 3D baselines.}
    \small
    \setlength\tabcolsep{4pt} 
    \begin{tabular}{l|ccccc}
    \toprule  
     & Ours & \multicolumn{2}{c}{R3DGS} & \multicolumn{2}{c}{ARGS} \\
    Sampling NO. & 40 & 40 & 256 & 40 & 256 \\
    \midrule
    Runtime  $\downarrow$ & \textbf{49} & 221 & 1480 & 1103 & 6318\\
    \midrule
    Relit PSNR $\uparrow$ & \textbf{27.977} & 19.214 & 21.983 & 21.336 & 23.879\\
    \bottomrule
    \end{tabular}
    \label{table:efficiency_comparison}
\end{table}
}
\efficiencycomparison

\begin{figure*}[t]
% \vspace{-10pt}
	\centering
	\includegraphics[width=\textwidth]{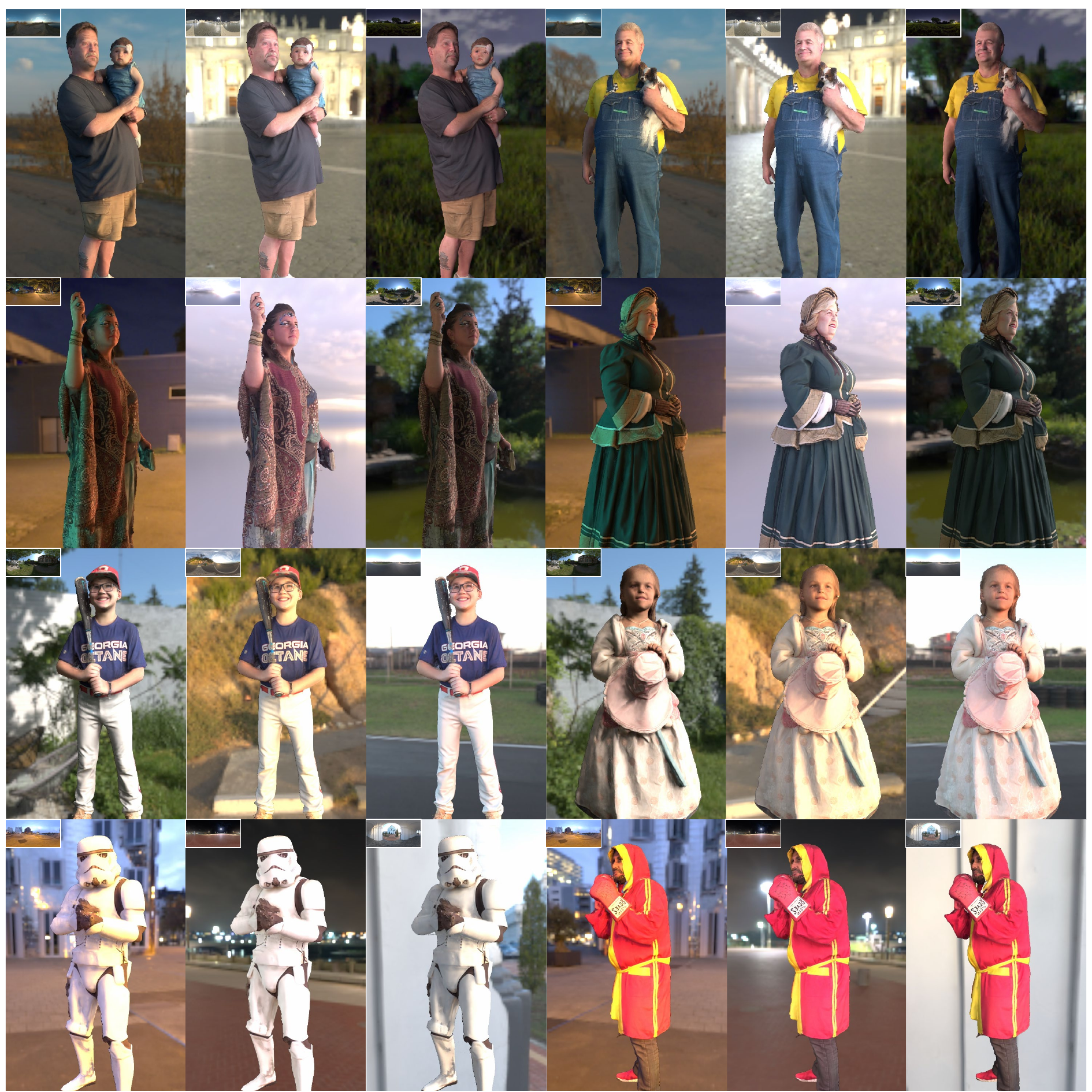}
	% \vspace{-15pt}
	\caption{Additional relighting results of various performers under diverse lighting conditions. Zoom in for the best view.}
	\label{fig:more_results}
 % \vspace{-15pt}
\end{figure*}

Furthermore, we conduct runtime and relighting quality comparisons against 3D baselines in Table~\ref{table:efficiency_comparison}, including R3DGS and ARGS. R3DGS accelerates rendering by precomputing visibility via ray tracing. 
To ensure a fair evaluation, we include this precomputation time in the total runtime. Specifically, we compare R3DGS and ARGS under two ray sampling configurations: 40 rays and 256 rays. The 256-ray setting is consistent with our main paper experiments, while the 40-ray setting is included to highlight efficiency under reduced sampling.
Both R3DGS and ARGS compute visibility using ray tracing over 3D Gaussian point clouds. ARGS suffers from significantly slower rendering speed due to its heavy Style-Unet architecture. 
However, our method achieves superior relighting quality while being significantly more computationally efficient due to the SH-Vis and light parametrization.

\paragraph{User study.}
We conducted a user study to evaluate the visual plausibility of our method on the proposed test set. In a pairwise comparison setup between our method and each baseline, 50 participants (including 40 students specializing in computer vision and graphics and 10 individuals from the general public) were asked to select the more visually convincing relighting result from each image pair. The study was conducted on 50 image samples. We report the preference rate (defined as the fraction of times participants favored our results over those of the baselines) in Table~\ref{tab:user}, where participants consistently rated our relighting outputs as more visually convincing and physically coherent than those of baseline methods.
\newcommand{\tableuser}{
\begin{table}[!htbp]
\caption{User preference rate.}
% \vspace{-2mm}
\small
\setlength\tabcolsep{5pt} 
\centering
\begin{tabular}{l|cccc}
\toprule 
Methods & R3DGS  & ARGS & IC-Light & SwitchLight \\ 
\midrule
Preference rate  & 0.93 & 0.85 & 0.77 & 0.65   \\ 

\bottomrule

\end{tabular}
% \vspace{-6mm}
\label{tab:user}
\end{table}
}
\tableuser
% \input{tables/nvs_depth}
% \nvsdepth

\section{Further Analysis}
\label{sec:extra experiment}
\begin{figure*}[t]
% \vspace{-10pt}
	\centering
	\includegraphics[width=\textwidth]{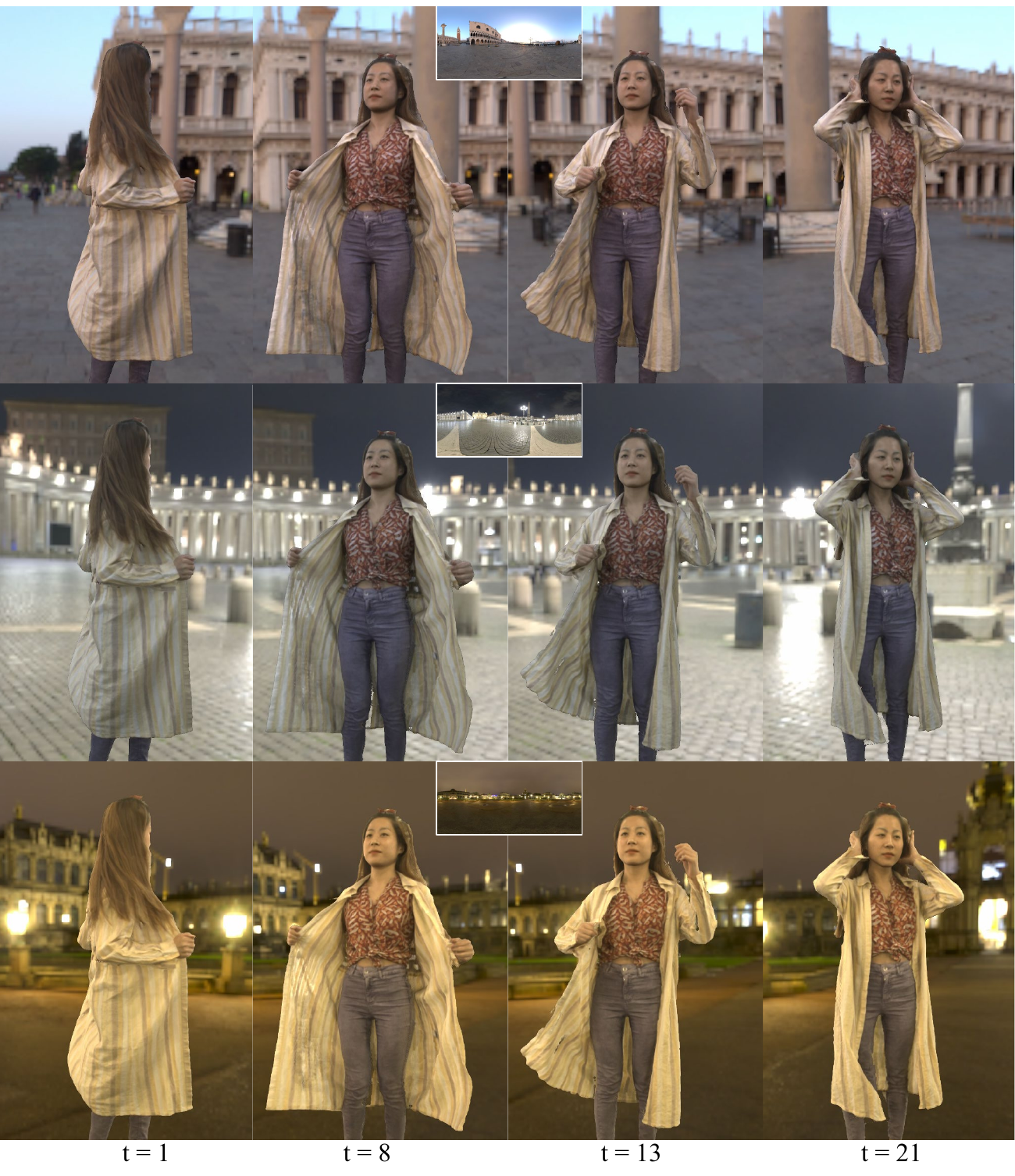}
	% \vspace{-15pt}
	\caption{Dynamic relighting case from GPS-Gaussian~\cite{zheng2024gps} under diverse lighting conditions and novel viewpoints.}
	\label{fig:dynamic_exp}
 % \vspace{-15pt}
\end{figure*}

\begin{figure*}[t]
% \vspace{-10pt}
	\centering
	\includegraphics[width=\textwidth]{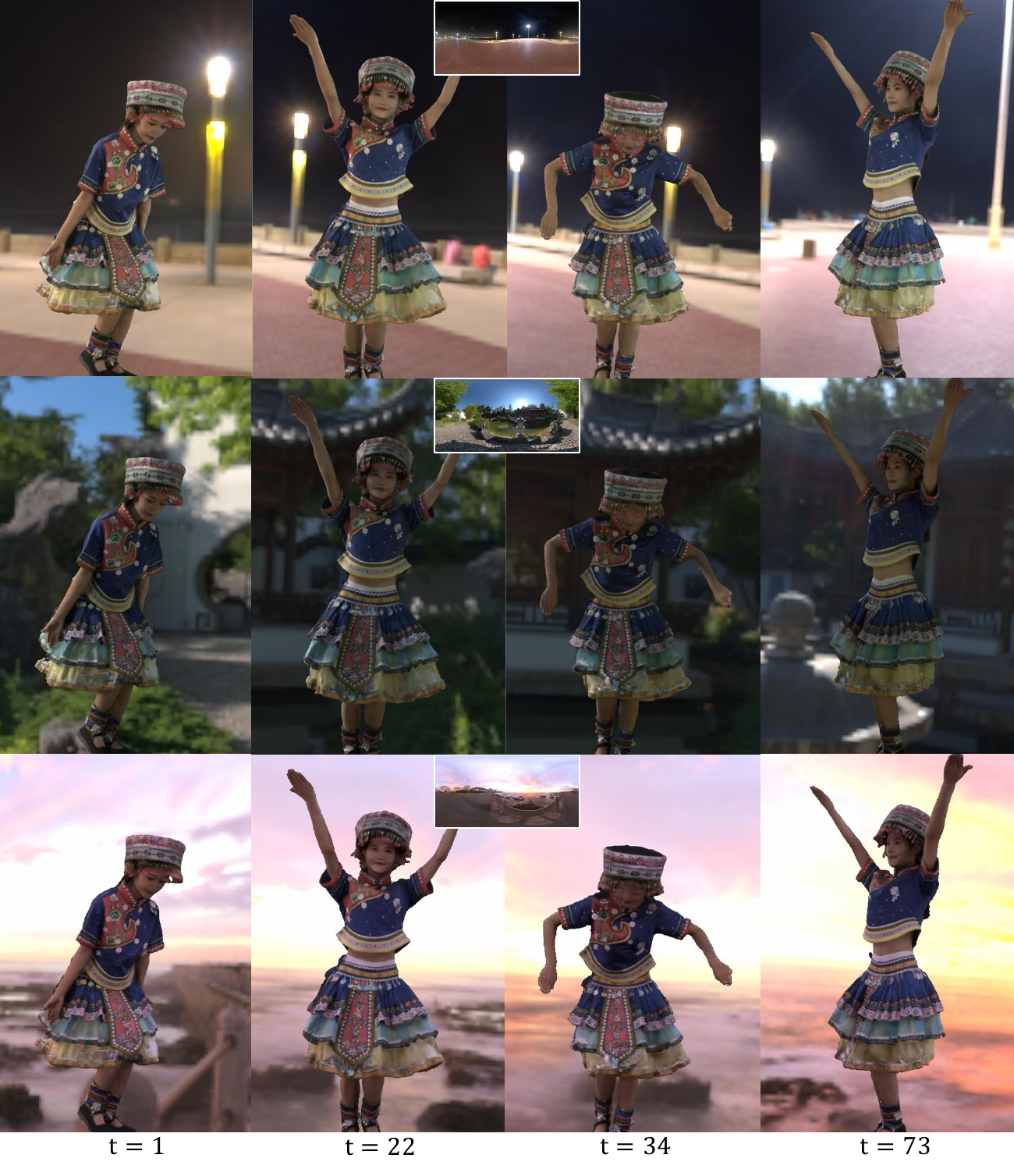}
	% \vspace{-15pt}
	\caption{Dynamic relighting case from DNA-Rendering~\cite{cheng2023dna} under diverse lighting conditions and novel viewpoints.}
	\label{fig:dynamic_exp2}
 % \vspace{-15pt}
\end{figure*}

\begin{figure*}[t]
% \vspace{-10pt}
	\centering
	\includegraphics[width=\textwidth]{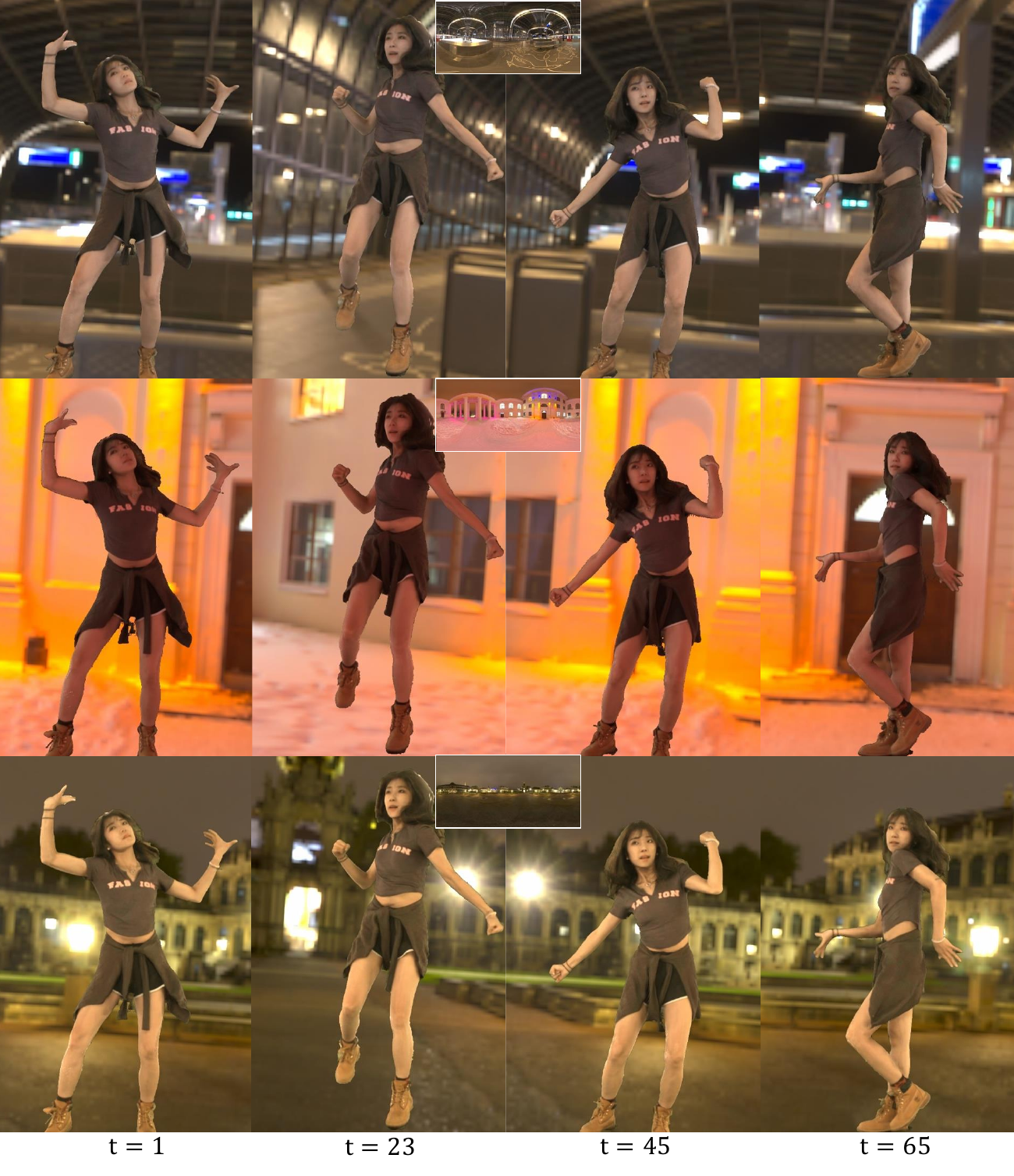}
	% \vspace{-15pt}
	\caption{Dynamic relighting case from DNA-Rendering~\cite{cheng2023dna} under diverse lighting conditions and novel viewpoints.}
	\label{fig:dynamic_exp3}
 % \vspace{-15pt}
\end{figure*}

\begin{figure*}[t]
% \vspace{-10pt}
	\centering
	\includegraphics[width=0.9\textwidth]{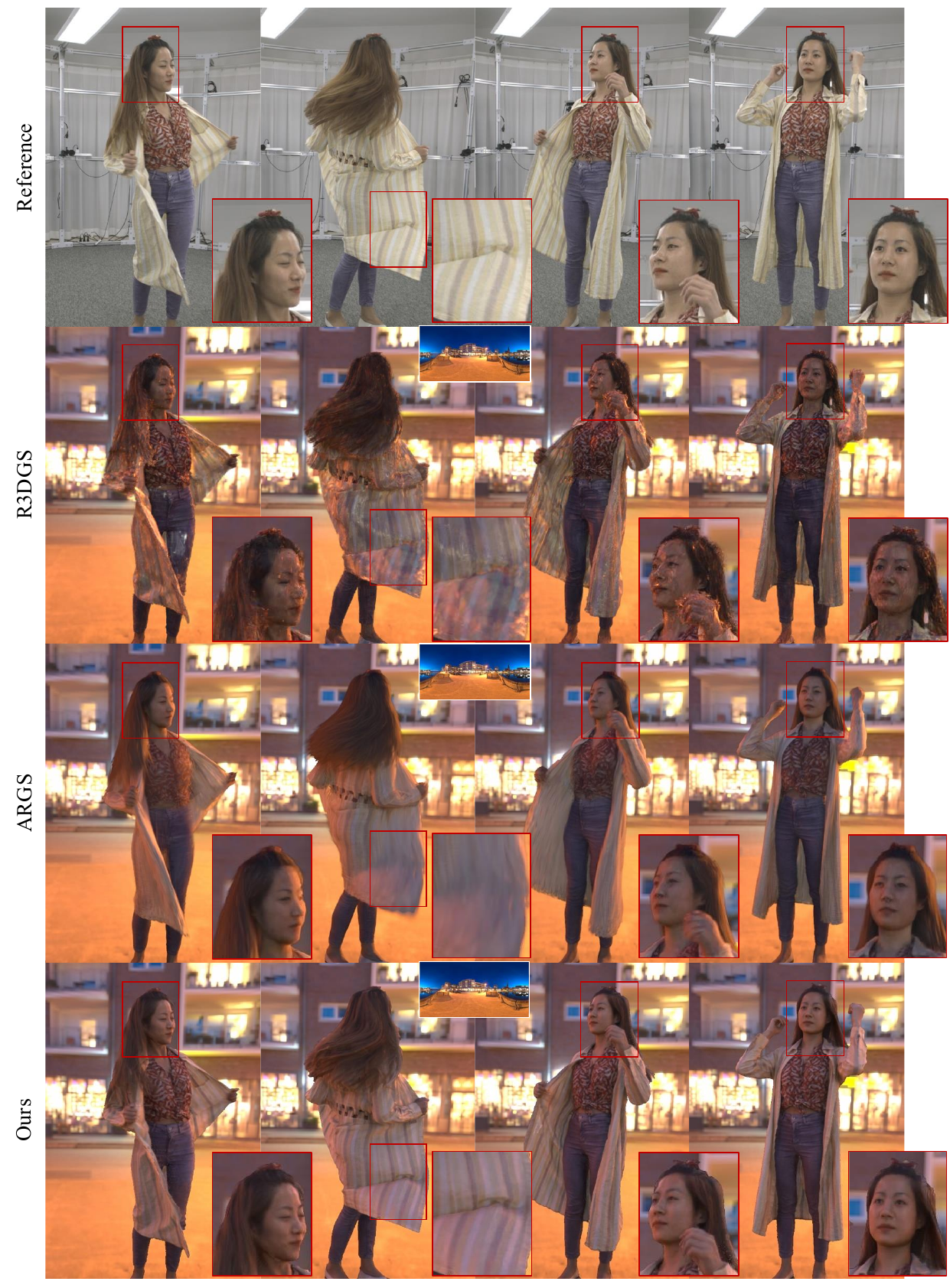}
	% \vspace{-15pt}
	\caption{Dynamic relighting comparison on the GPS-Gaussian~\cite{zheng2024gps} dataset.}
	\label{fig:dynamic_comparison1}
 % \vspace{-15pt}
\end{figure*}

\begin{figure*}[t]
% \vspace{-10pt}
	\centering
	\includegraphics[width=0.9\textwidth]{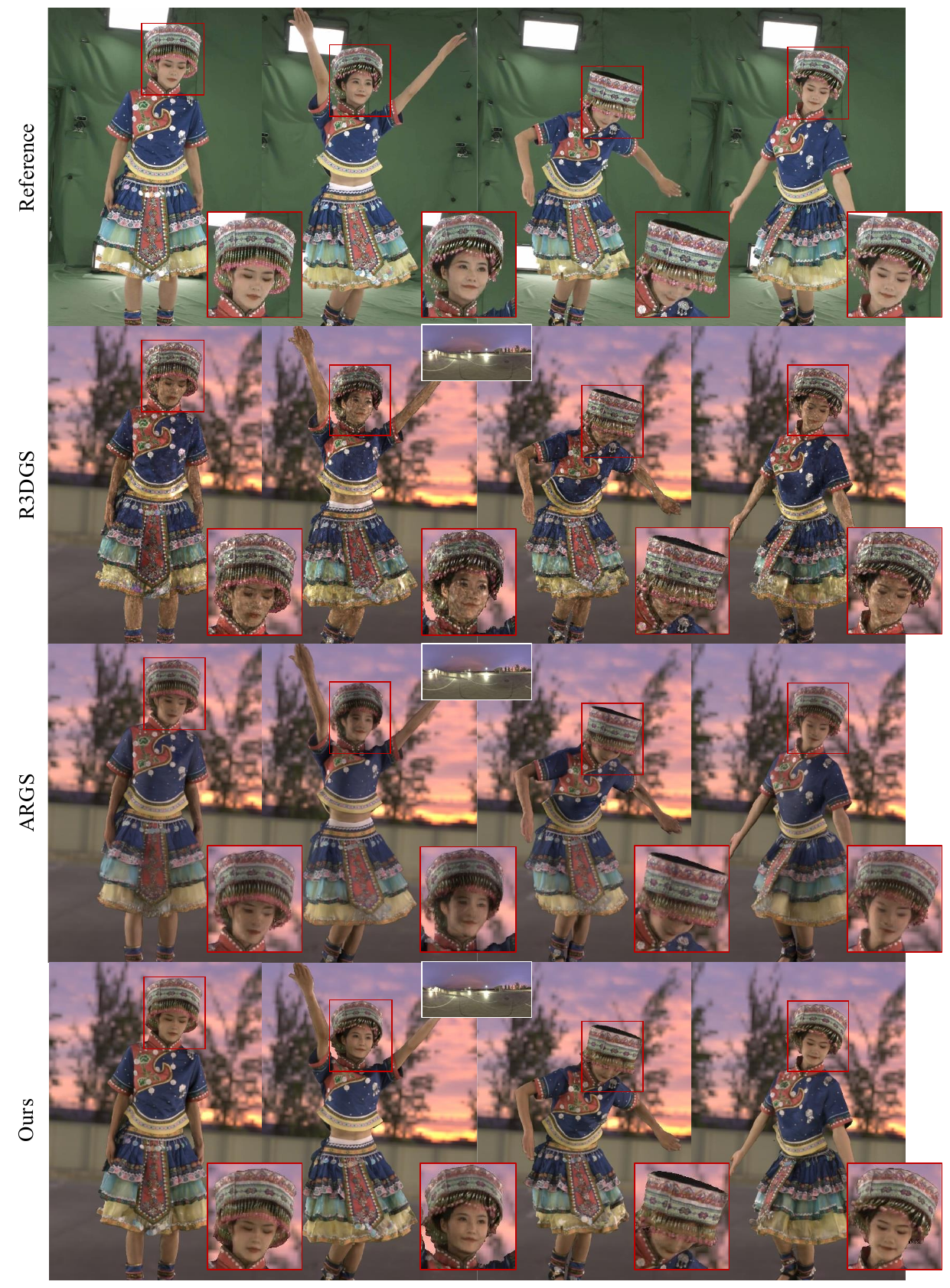}
	% \vspace{-15pt}
	\caption{Dynamic relighting comparison on the DNA-Rendering~\cite{cheng2023dna} dataset.}
	\label{fig:dynamic_comparison2}
 % \vspace{-15pt}
\end{figure*}

% \paragraph{Implementation details.}
\paragraph{Geometry reconstruction.}
% Table~\ref{tab:lighting_prior} compares novel view synthesis (NVS) and depth accuracy between GPS-Gaussian~\cite{zheng2024gps} and our method, highlighting the effectiveness of our lighting-invariant priors in enhancing both geometric reconstruction and appearance fidelity. 
% %
% Note that we evaluate depth accuracy using End-Point Error (EPE) and the 1-pixel accuracy metric, which measure the average disparity error and the percentage of pixels with depth error less than one pixel, respectively.
Table~\ref{table:mae} demonstrates the effectiveness of the proposed normal refinement module by evaluating surface normal quality using the MAE metric. While the incorporation of lighting-robust features facilitates the recovery of smooth normals, fine-grained geometric details remain lacking due to the limitations of convex upsampling. In contrast, our normal refinement module successfully captures high-frequency details, leading to improvements in MAE.
\newcommand{\nml}{
\begin{table}[!htbp]
    \centering
    \caption{MAE Comparison}
    \small
    \setlength\tabcolsep{3pt} 
    \begin{tabular}{l|ccc}
    \toprule  
    Method & GPS-Gaussian~\cite{zheng2024gps} & Coarse Normal & Refined Normal \\
    \midrule
    MAE $\downarrow$ & 8.145 & 5.903 & \textbf{5.369}\\
    \bottomrule
    \end{tabular}
    \label{table:mae}
\end{table}
}
\nml
\paragraph{Inference speed}
The end-to-end pipeline, including geometry reconstruction, material prediction, and relighting, runs at approximately 20 FPS on an RTX 4090 GPU. 
%Specifically, once the geometry has been reconstructed and material parameters have been predicted, the rendering part alone runs at ~60 FPS. 
The runtime breakdown of each module is shown in Table~\ref{tab:runtime}.
\paragraph{More results.}
We present additional relighting results of various performers under diverse lighting conditions in Fig.~\ref{fig:more_results}. The photorealistic quality, lighting consistency, and physical plausibility observed in these results highlight the effectiveness and generalization of our proposed method.
\newcommand{\runtime}{
\begin{table}[t]
\caption{Run time breakdown of each module.}
% \vspace{-2mm}
\small
\setlength\tabcolsep{2pt} 
\centering
\begin{tabular}{l|cccc}
\toprule 
Methods & Geometry  & Gaussian attributes & Rendering & Total \\ 
\midrule
Run time (ms)  & 11 & 21 & 17 & 49   \\ 

\bottomrule

\end{tabular}
% \vspace{-6mm}
\label{tab:runtime}
\end{table}
}
\runtime
\paragraph{Dynamic extension.}
Although GRGS is not explicitly designed with a temporal consistency module for dynamic scenarios, it can be extended to dynamic settings under uniform lighting conditions, as shown in ~\cref{fig:dynamic_exp},~\cref{fig:dynamic_exp2} and ~\cref{fig:dynamic_exp3}. By treating input images as diffuse albedo, the method mitigates most flickering artifacts caused by temporal inconsistencies. We further provide a qualitative comparison on dynamic sequences against R3DGS~\cite{gao2024relightable} and ARGS~\cite{li2024animatable}, as shown in Fig.~\ref{fig:dynamic_comparison1} and Fig.~\ref{fig:dynamic_comparison2}.
R3DGS struggles to reconstruct geometry from sparse-view inputs, resulting in severe artifacts in relighting.
ARGS alleviates this issue by employing a body template, which improves geometric consistency. However, its template deformation capability is limited, making it difficult to handle loose or highly deformable clothing such as long dresses.
Moreover, both approaches require extensive optimization, taking 3–4 days to process a 200-frame sequence, which restricts practicality.

In contrast, our method requires no extra optimization, achieves near real-time inference, and does not rely on body templates, enabling robust generalization to diverse human identities and clothing styles.
With the proposed PGNR module, our approach further delivers more natural, coherent, and physically plausible relighting results across dynamic sequences.

\section{Loss Functions}
\label{sec:optimization}
\paragraph{LGR loss.} We supervise geometry reconstruction using a depth loss $\mathcal{L}_{\text{depth}}$ and a normal loss $\mathcal{L}_{\text{normal}}$, formulated as:
\begin{equation}
    \mathcal{L}_{\text{LGR}} = \mathcal{L}_{\text{depth}} + \mathcal{L}_{\text{normal}}
\end{equation}
The depth loss $\mathcal{L}_{\text{depth}}$ is defined as the weighted L1 distance between the predicted depth sequence $\{\mathbf{d}_1, \cdots, \mathbf{d}_N\}$ and ground truth depth $\mathbf{d}_{\text{GT}}$ with exponentially increasing weights following~\cite{lipson2021raft}:
\begin{equation}
    \mathcal{L}_{\text{depth}} = \sum_{i=1}^N \mu^{N-i}||\mathbf{d}_i - \mathbf{d}_{\text{GT}}||_1
\end{equation}
where $\mu$ is set to 0.9. For the normal loss, given ground truth $\mathbf{N}_{\text{GT}}$, we combine the L1 distance and a perceptual loss~\cite{zhang2018unreasonable} to ensure both geometric accuracy and surface smoothness:
\begin{equation}
    \mathcal{L}_{\text{normal}} = \mathcal{L}_1(\mathbf{N}_f, \mathbf{N}_{\text{GT}}) + \lambda_1\mathcal{L}_{\text{percep}}(\mathbf{N}_f, \mathbf{N}_{\text{GT}})
\end{equation}

\paragraph{PGNR loss.} We define the overall loss as a combination of $\mathcal{L}_{\text{albedo}}$, material smoothness loss $\mathcal{L}_{\text{smooth}}$, light transport loss $\mathcal{L}_{\text{LT}}$, and PBR loss $\mathcal{L}_{\text{PBR}}$ to facilitate appearance reconstruction and relighting via 2D-to-3D supervision:
\begin{equation}
    \mathcal{L}_{\text{PGNR}} = \mathcal{L}_{\text{albedo}} + \mathcal{L}_{\text{smooth}} +  \mathcal{L}_{\text{LT}}+ \mathcal{L}_{\text{PBR}}
\end{equation}
Note that after PBR and rasterization, the material maps: $\mathbf{A} \ \text{and} \ \boldsymbol{\Gamma}$; the visibility $V$ parameterized by SH map $\mathbf{Y}_{\text{vis}}$; the indirec light $\mathbf{L}_{\text{ind}}$ parameterized by SH map $\mathbf{Y}_{\text{ind}}$; the direct light $\mathbf{L}_{\text{d}}$, and the PBR color $\mathbf{C}_{\text{PBR}}$ are all rendered into the 2D image space as $\mathbf{I}_{\text{albedo}}$, $\mathbf{I}_{\text{rough}}$, $\mathbf{I}_{\text{ao}}$, $\mathbf{I}_{\text{indl}}$, $\mathbf{I}_{\text{dl}}$ and $\mathbf{I}_{\text{PBR}}$, respectively. Thus, they can be supervised via corresponding 2D ground truths.
Specifically, the albedo loss is composed of an L1 loss and a perceptual loss to measure the difference between the predicted albedo and the ground truth $\mathbf{I}_{\text{albedo}}^\prime$:
\begin{equation}
    \mathcal{L}_{\text{albedo}} = \mathcal{L}_1(\mathbf{I}_{\text{albedo}}, \mathbf{I}_{\text{albedo}}^\prime) + \lambda_2\mathcal{L}_{\text{percep}}(\mathbf{I}_{\text{albedo}}, \mathbf{I}_{\text{albedo}}^\prime)
\end{equation}
To promote spatially smooth material estimation, we adopt a bilateral smoothness loss following~\cite{gao2024relightable}:
% \begin{equation}
%     \mathcal{L}_{\text{smooth}} = \lambda_3||\nabla \mathbf{\mathbf{I}_{\text{albedo}}}||\text{exp}\left(-||\nabla \mathbf{I}_{\text{albedo}}^\prime||\right) + \lambda_4||\nabla \mathbf{\mathbf{I}_{\text{rough}}}||\text{exp}\left(-||\nabla \mathbf{I}_{\text{albedo}}^\prime||\right)
% \end{equation}
\begin{equation}
\begin{split}
    \mathcal{L}_{\text{smooth}} ={}& 
    \lambda_3 \|\nabla \mathbf{I}_{\text{albedo}}\|
    \exp\left(-\|\nabla \mathbf{I}_{\text{albedo}}^\prime\|\right) \\
    &+ \lambda_4 \|\nabla \mathbf{I}_{\text{rough}}\|
    \exp\left(-\|\nabla \mathbf{I}_{\text{albedo}}^\prime\|\right)
\end{split}
\end{equation}
Note that ground truth roughness maps are not available; thus, the roughness map $\boldsymbol{\Gamma}$ is learned implicitly through the entire PBR framework.
The light transport loss is combined with ambient occlusion loss $\mathcal{L}_{\text{ao}}$, direct light loss $\mathcal{L}_{\text{d}}$, and indirect light loss $\mathcal{L}_{\text{ind}}$, we use L1 distance to measure the predicted and ground truth ones:
\begin{equation}
    \mathcal{L}_{\text{LT}} = \mathcal{L}_1(\mathbf{I}_{\text{ao}}, \mathbf{I}_{\text{ao}}^\prime) + \mathcal{L}_1(\mathbf{I}_{\text{dl}}, \mathbf{I}_{\text{dl}}^\prime) +
    \mathcal{L}_1(\mathbf{I}_{\text{indl}}, \mathbf{I}_{\text{indl}}^\prime)
\end{equation}
The PBR loss applies an L1 loss and a perceptual loss for measuring the overall relighting quality:
\begin{equation}
    \mathcal{L}_{\text{PBR}} = \mathcal{L}_1(\mathbf{I}_{\text{PBR}}, \mathbf{I}_{\text{PBR}}^\prime) + \lambda_{5}\mathcal{L}_1(\mathbf{I}_{\text{PBR}}, \mathbf{I}_{\text{PBR}}^\prime)
\end{equation}
The loss weights are set as follows: $\lambda_1 = 0.2$, $\lambda_2 = 0.2$, $\lambda_3 = 0.1$, $\lambda_4 = 0.1$, and $\lambda_5 = 0.2$.
\section{Limitations}
\label{sec:limitations}
% While our method delivers high-quality relighting results on human subjects, several challenging cases remain. Due to the inherent limitations of the adopted rendering equation, our approach is unable to accurately model transparent materials such as eyeglasses and struggles to handle extremely thin structures, including hair strands. Future endeavors could explore the implementation of advanced reflectance models to further enhance the realism of relighting.
While our method achieves high-quality relighting results on human subjects, it has two primary limitations. First, due to the intrinsic constraints of the adopted rendering equation, our approach cannot accurately model transparent or metallic materials, such as eyeglasses and metallic sequins, and struggles with extremely thin structures, including individual hair strands. Future work could explore the integration of advanced reflectance models to further improve the realism of relighting.

Second, our method requires at least two input images with precise camera poses for stereo-based geometry estimation and relighting. This dependency on multi-view calibration limits its applicability in scenarios with unknown or imprecise poses and reduces robustness in unconstrained environments. In future work, we aim to incorporate geometry priors from foundational models, such as VGGT~\cite{wang2025vggt}, enabling novel view synthesis and high-quality relighting from arbitrary viewpoints without explicit calibration.

% {
%     \small
%     \bibliographystyle{ieeenat_fullname}
%     \bibliography{main}
% }
\end{document}